\documentclass{article}

\usepackage{microtype}
\usepackage{graphicx}

\usepackage{subcaption} 
\usepackage{caption}

\usepackage{booktabs} 

\usepackage{hyperref}

\usepackage[accepted]{icml2023}

\usepackage{amsmath}
\usepackage{amssymb}
\usepackage{mathtools}
\usepackage{amsthm}

\usepackage[capitalize,noabbrev]{cleveref}

\theoremstyle{plain}
\newtheorem{theorem}{Theorem}[section]
\newtheorem{proposition}[theorem]{Proposition}

\theoremstyle{definition}

\theoremstyle{remark}

\usepackage[textsize=tiny]{todonotes}

\newcommand{\E} {{\mathbb E}}

\newcommand{\B}{\mathbf}
\newcommand{\Bs}{\boldsymbol}
\newcommand{\om} {{\omega}}
 
\DeclareMathOperator*{\argmin}{arg\,min}
\usepackage{enumitem}
\icmltitlerunning{Unsupervised Source Separation with Limited Data}

\begin{document}

\twocolumn[\icmltitle{Unearthing InSights into Mars:\\
Unsupervised Source Separation with Limited Data}




\begin{icmlauthorlist}
\icmlauthor{Ali Siahkoohi}{1}
\icmlauthor{Rudy Morel}{2}
\icmlauthor{Maarten V. de Hoop}{1}
\icmlauthor{Erwan Allys}{3}
\icmlauthor{Grégory Sainton}{4}
\icmlauthor{Taichi Kawamura}{4}
\end{icmlauthorlist}

\icmlaffiliation{1}{Department of Computational Applied Mathematics \& Operations Research, Rice University}
\icmlaffiliation{2}{D\'{e}partement d'informatique de l'ENS, ENS, CNRS, PSL University, Paris, France}
\icmlaffiliation{3}{Laboratoire de Physique de l'\'Ecole normale sup\'erieure, ENS, Universit\'e PSL, CNRS, Sorbonne Universit\'e, Universit\'e Paris Cit\'e, F-75005 Paris, France}
\icmlaffiliation{4}{Institut de Physique du Globe de Paris}

\icmlcorrespondingauthor{Ali Siahkoohi}{alisk@rice.edu}

\icmlkeywords{Source Separation, Unsupervised Learning, Inductive Bias,
Time Series Analysis, Signal Processing, Scattering Networks, Scattering
Covariances, Machine Learning, ICML}

\vskip 0.3in ]



\printAffiliationsAndNotice{}

\begin{abstract}
Source separation involves the ill-posed problem of retrieving a set of source signals that have been observed through a mixing operator. Solving this problem requires prior knowledge, which is commonly incorporated by imposing regularity conditions on the source signals, or implicitly learned through supervised or unsupervised methods from existing data. While data-driven methods have shown great promise in source separation, they often require large amounts of data, which rarely exists in planetary space missions. To address this challenge, we propose an unsupervised source separation scheme for domains with limited data access that involves solving an optimization problem in the wavelet scattering covariance representation space---an interpretable, low-dimensional representation of stationary processes. We present a real-data example in which we remove transient, thermally-induced microtilts---known as glitches---from data recorded by a seismometer during NASA's InSight mission on Mars. Thanks to the wavelet scattering covariances' ability to capture non-Gaussian properties of stochastic processes, we are able to separate glitches using only a few glitch-free data snippets.
\end{abstract}

\section{Introduction}
\label{intro}

\begin{figure}[!t]
    \centering

    \begin{subfigure}[b]{1.0\linewidth}
        \includegraphics[width=1.0\linewidth]{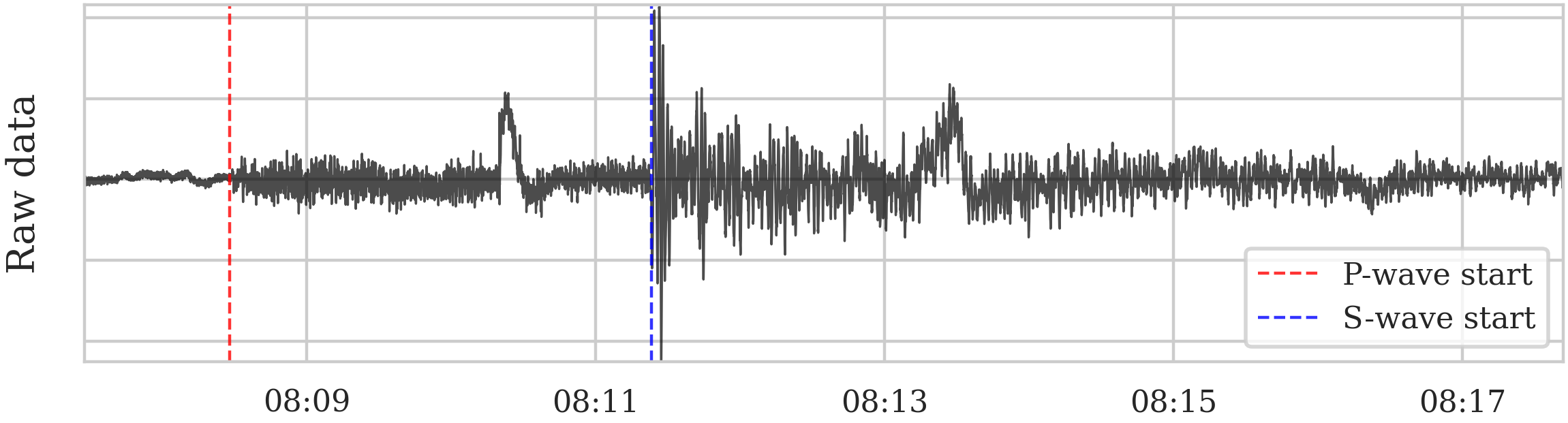}
    \end{subfigure}\hspace{0em}

    \begin{subfigure}[b]{1.0\linewidth}
        \includegraphics[width=1.0\linewidth]{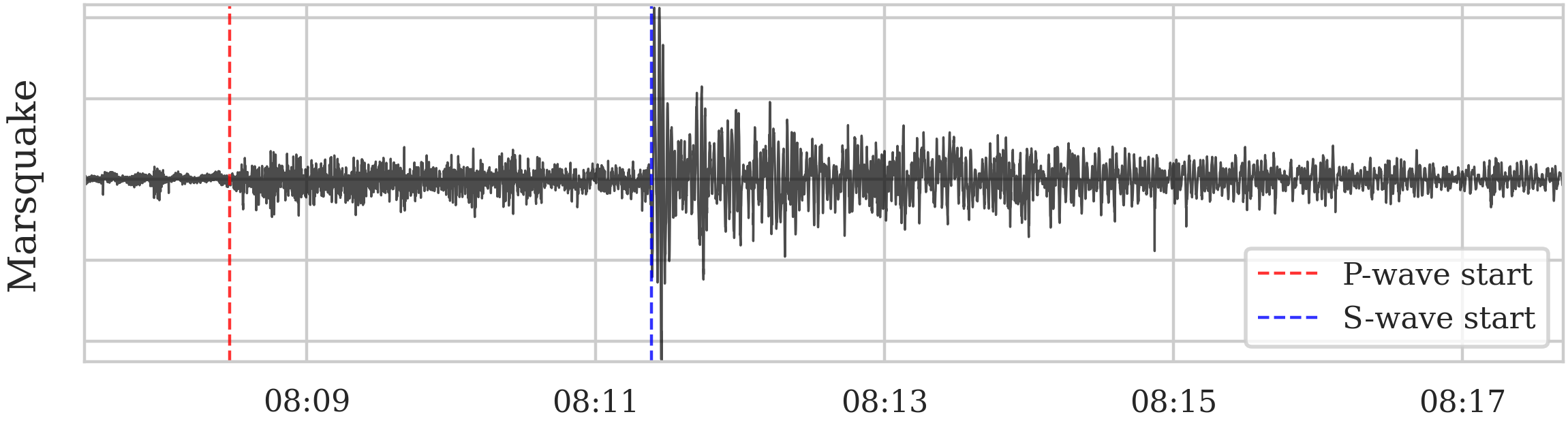}
    \end{subfigure}\hspace{0em}

    \begin{subfigure}[b]{1.0\linewidth}
        \includegraphics[width=1.0\linewidth]{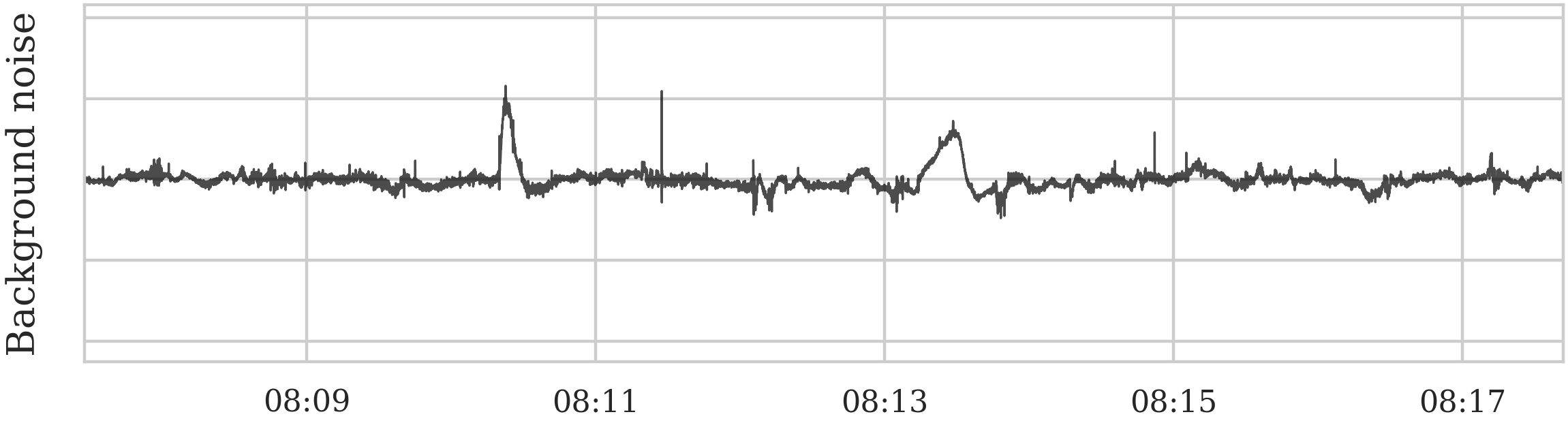}
    \end{subfigure}\hspace{0em}
    \caption{Unsupervised separation of background noise, including thermally induced microtilts (glitches), from a marsquake recorded by the InSight lander's seismometer on February 3, 2022~\cite{doi.org/10.12686/a19}. Approximately 30 hours of raw data from the U component, with no recorded marsquakes, were utilized for background noise separation without any explicit prior knowledge of marsquakes or glitches. The horizontal axis represents the UTC time zone.}
    \label{fig:marsquake_cleaning}
\end{figure}

Source separation is a problem of fundamental importance in the field of signal processing, with a wide range of applications in various domains such as telecommunications~\cite{CHEVREUIL2014135, gay2012acoustic, KHOSRAVY2020423}, speech processing~\cite{Pedersen2008, 7602895, 6854299}, biomedical signal processing~\cite{7214354, 10.1117/12.481361, 8303715} and geophysical data processing~\cite{doi:10.1190/geo2013-0168.1, doi:10.1190/geo2015-0108.1, 10.1029/2020EA001317}. Source separation arises when multiple source signals of interest are combined through a mixing operator. The goal is to estimate the original sources with minimal prior knowledge of the mixing process or the source signals themselves. This makes source separation a challenging problem, as the number of sources is usually unknown, and the sources are often non-Gaussian, nonstationary, and multiscale.

Classical signal-processing based source separation
methods~\cite{266878, JUTTEN19911, doi:10.1142/S0129065700000028,
544786, 720250, JUTTEN2004217} while being extensively studied and well
understood, often make simplifying assumptions regarding the sources, e.g., sources being distributed according to Gaussian or Laplace distributions,
which might negatively bias the outcome of source
separation~\cite{720250,10.5555/945365.964305}. To partially address the
shortcomings of classical approaches, deep learning methods have been
proposed as an alternative approach for source separation, which exploit
the information in existing datasets to learn prior information about
the sources. In particular, supervised learning methods
\cite{jang2003maximum, 10.1109/ICASSP.2016.7471631,
10.1007/s11042-020-09419-y, 8802366, wang2018supervised} commonly rely
on existence of labeled training data and perform source separation
using an end-to-end training scheme. However, since they require access
to ground truth source signals for training, supervised methods are
limited to domains in which labeled training data is available.

On the other hand, unsupervised source separation methods
\cite{10.1162/neco.2008.04-08-771, drude2019unsupervised,
NEURIPS2020_28538c39, LIU2022100616, 9747202, 9616154} do not rely on
the existence of labeled training data and instead attempt to infer the
sources based on the properties of the observed signals. These methods
make minimal assumptions about the underlying sources, which make them a
suitable choice for realistic source separation problems. Despite their
success, unsupervised source separation methods often require tremendous
amount of data during training \cite{NEURIPS2020_28538c39}, which is
often infeasible in certain applications such as problem arising in
planetary space missions, e.g., due to challenges associated with
data acquisition. Moreover, generalization concerns preclude the use of
data-driven methods trained on synthetic data in real-world applications
due to the discrepancies between synthetic and real data.

To address these challenges, we propose an unsupervised source
separation method applicable to domains with limited access to data. In
order to achieve this, we embed inductive biases into our approach
through the use of domain knowledge from time-series analysis and signal
processing via the an extension of scattering networks~\cite{bruna2013invariant}. As a
means of capturing non-Gaussian and multiscale characteristics of the
sources, we extract second-order information of scattering coefficients,
known as the wavelet scattering covariance
representation~\cite{morel2022scale}. We perform source separation by
solving an optimization problem over the unknown sources that entails
minimizing multiple carefully selected and normalized loss functions in
the wavelet scattering covariance representations space. These loss
function are designed to: (1) ensure data-fidelity, i.e., enforce the
recovered sources to explain the observed (mixed) data; (2) incorporate
prior knowledge in the form of limited (e.g., $\approx 50$) training
examples from one of the sources; and (3) impose a notion of statistical
independence between the recovered sources. Our proposed method does not
require any labeled training data, and can effectively separate sources
even in scenarios where access to data is limited.

As a motivating example, we apply our approach to data recorded by a
seismometer on Mars during NASA's Interior Exploration using Seismic
Investigations, Geodesy and Heat Transport (InSight)
mission~\cite{giardini2020seismicity, golombek2020geology,
knapmeyer2020nasa}. The InSight lander's seismometer---known as the SEIS instrument---detected
marsquakes~\cite{10.1785/0320220007, CEYLAN2022106943,
10.1029/2022GL101270, doi.org/10.12686/a19} and transient atmospheric
signals, such as wind and temperature changes, that provide information
about the Martian atmosphere~\cite{10.1093/gji/ggac464} and enable
studying the interior structure and composition of the Red
Planet~\cite{10.1029/2022GL101508}. The signal recorded by the InSight
seismometer is heavily influenced by atmospheric activity and surface
temperature~\cite{lognonne2020constraints, LORENZ2021114119}, resulting
in a distinct daily pattern. Among different types of noise, transient thermally induced microtilts,
commonly referred to as glitches~\cite{10.1029/2020EA001317,
barkaoui2021anatomy}, are a significant component of the noise and one
of the most frequent recorded events. These glitches, hinder the
downstream analysis of the data if left
uncorrected~\cite{10.1029/2020EA001317}. We show that our method is
capable of removing glitches from the recorded data by only using a few
snippets of glitch-free data.

In the following sections, after describing the related work, we introduce wavelet scattering covariance
as a domain-knowledge rich representation for analyzing time-series and
provide justification for their usage in the context of source
separation. As a means to perform source separation in domains with
limited data, we introduce our source separation approach that involves
solving an optimization problem with loss functions defined in the
wavelet scattering covariance space. We present two numerical
experiments: (1) a synthetic setup in which we can quantify the accuracy
of our method; and (2) examples involving seismic data recorded during the NASA InSight mission.

\section{Related Work}
\label{related}

\citet{regaldo2021new} introduced the notion of components separation
through a gradient descent in signal space with indirect constraints
with applications to to the separation of an astrophysical emission
(polarized dust emission in microwave) and instrumental noise. In an
extensive study,~\citet{refId0} attempts to separate the full sky
observation of the dust emission with instrumental noise using similar
techniques via wavelet scattering covariance representations. Authors
take the nonstationarity of the signal into account by constraining
statistics on several sky masks. Contrarily to a usual denoising
approach, both of these works focus primarily on recovering the
statistics of the signal of interest. In a related
approach,~\citet{jeffrey2022single} use a scattering transform
generative model to perform source separation in a Bayesian framework.
While very efficient, this approach requires training samples from each
component, which are often not available. Finally,
\citet{10.1785/0220210361} similarly aim to remove glitches and they
develop a supervised learning based on deglitched data obtained by
existing glitch removal tools. As a result, the accuracy of their result
is limited to the accuracy of the underlying data processing tool, which
our method avoid by being unsupervised. As we show in our examples, we
are able to detect and remove glitches that were undetected by the main
deglitching software~\cite{10.1029/2020EA001317} developed closely by
the InSight team.

\section{Wavelet Scattering Covariance}
\label{scatcov}

In order to enable unsupervised source separation with limited
quantities of data, we propose to design a low-dimensional,
domain-knowledge rich representation of data with which we perform
source separation. This is partially motivated by recent success of
self-supervised learning methods in natural language processing where
high-performing representations of data---obtained through pretrained
Transformers~\cite{vaswani2017attention, baevski2020wav2vec,
gulati2020conformer, zhang2020pushing}---are used in place of raw data
to successfully perform various downstream tasks~\cite{polyak2021speech,
gulati2020conformer, baevski2020wav2vec, zhang2020pushing, chung2021w2v,
interspeech22}.

Due to our limited access to data, we cannot employ self-supervised learning with Transformers to acquire
high-performing data representations. Instead, we propose to use wavelet scattering
covariances~\cite{morel2022scale} as means to transfer data to a
suitable representation space for source separation. Rooted in scattering networks~\cite{bruna2013invariant}, wavelet scattering covariances provide
interpretable representations of data and are able to characterize a wide
range of non-Gaussian properties of multiscale stochastic
processes~\cite{morel2022scale}---a type of signals that we consider
in this paper. The wavelet scattering covariance generally does not
require any pretraining and its weights, i.e., wavelets in the
scattering network, are often chosen beforehand (see~\citet{Seydoux2020}
for a data-driven wavelet choice) according to the time-frequency
properties of data. In the next section, we introduce the construction
of this representation space by first describing scattering networks.

\subsection{Wavelet Transform and Scattering Networks}

The main ingredient of the wavelet scattering covariance representation
is a scattering network~\cite{bruna2013invariant} that consists of a
cascade of wavelet transforms followed by a nonlinear activation
function (akin to a typical convolutional neural network). In this
network architecture, the wavelet transform, denoted by a linear
operator $\B{W}$, is a convolutional operator with predefined kernels,
i.e., wavelet filters. These filters include a low-pass filter
$\varphi_J(t)$ and $J$ complex-valued band-pass filters
$\psi_j(t)=2^{-j}\psi(2^{-j}t),\ 1\leq j\leq J$, which are obtained by
the dilation of a mother wavelet $\psi(t)$ and have zero-mean and a fast
decay away from $t=0$. The wavelet transform is often followed by the
modulus operator in scattering networks. The output of a two-layer
scattering network $S$ can be written as,%
\begin{equation}
 S(\B{x}) := \begin{bmatrix}
\B{W}\B{x}\\
\B{W}|\B{W}\B{x}|
\end{bmatrix},
\end{equation}
where $\B{W}\B{x}:=\B{x} \star \psi_j(t)$ denotes the wavelet transform
that extracts variations of the input signal $\B{x}(t)$ around time $t$
at scale $2^j$, and $|\cdot|$ is the modulus activation
function~\cite{bruna2013invariant}. The second component
$\B{W}|\B{W}\B{x}|$ computes the variations at different time and scales
of the wavelet coefficients $\B{W}\B{x}$.
The scattering transform yields features that characterize time
evolution of signal envelopes at different scales. Even though such
representation has many successful applications, e.g., intermittency
analysis \cite{bruna2015intermittent}, clustering \cite{Seydoux2020},
event detection and segmentation \cite{rodriguez2021recurrent} (with
learnable wavelets), it is not sufficient to build accurate models of
multiscale processes as it does not capture crucial dependencies across
different scales~\cite{morel2022scale}.

\subsection{Capturing Non-Gaussian Characteristics of Stochastic Processes}
\label{scatcov_formulation}

The dependencies across different scales in scattering transform
coefficients are crucial in characterizing and discriminating
non-Gaussian signals~\cite{morel2022scale}. To capture them, we explore
the outer product of the scattering coefficients matrix
$S(\B{x})S(\B{x})^\text{H}$:
\begin{equation}
\begin{bmatrix}
\B{W}\B{x} \left(\B{W}\B{x}\right)^\text{H} & \B{W}\B{x} \left(\B{W}|\B{W}\B{x}|\right)^\text{H}\\
\B{W}|\B{W}\B{x}| \left(\B{W}\B{x}\right)^\text{H} & \B{W}|\B{W}\B{x}| \left(\B{W}|\B{W}\B{x}|\right)^\text{H}
\end{bmatrix}.
\end{equation}
In the above expression, $^\text{H}$ denotes the conjugate transpose operation. The above matrix contains three types of coefficients:
\begin{itemize}[noitemsep]
    \item The correlation coefficients $\B{W}\B{x}
    \left(\B{W}\B{x}\right)^\text{H}$ across scales form a quasi-diagonal
    matrix, because separate scales do not correlate due to phase
    fluctuation, whether separate scales are dependent or not~\cite{morel2022scale}. We thus only keep its diagonal
    coefficients, which correspond to the wavelet power spectrum;
    \item The correlation coefficients $\B{W}\B{x}
    \left(\B{W}|\B{W}\B{x}|\right)^\text{H}$ capture signed interaction
    between wavelet coefficients. In particular, they detect
    sign-asymmetry and time-asymmetry in $\B{x}$~\cite{morel2022scale}.
    We also consider a diagonal approximation to this matrix. For the
    same reason as $\B{W}\B{x} \left(\B{W}\B{x}\right)^\text{H}$,
    two separate scales on the last wavelet operator do not correlate. However, there may exist a correlation between $\B{W}\B{x}$ at a given scale and $|\B{W}\B{x}|$ at a distinct scale, and we retain these;
    \item Finally coefficients $\B{W}|\B{W}\B{x}|
    \left(\B{W}|\B{W}\B{x}|\right)^\text{H}$ capture correlations between signal envelopes $|\B{W}\B{x}|$ at different scales. These correlations account for intermittency and time-asymmetry~\cite{morel2022scale}. Once again, we retain only those coefficients that demonstrate correlation between same-scale channels on the second wavelet operator.
\end{itemize}
We denote $\operatorname{diag}\left(S(\B{x})S(\B{x})^\text{H}\right)$ as
such diagonal approximation of the full sparse matrix
$S(\B{x})S(\B{x})^\text{H}$. The \emph{wavelet scattering covariance}
representation is obtained by computing the time average (average pool,
denoted by $\operatorname{Ave}$) of this diagonal approximation:
\begin{equation}
    \label{ScatteringCovariance}
    \Phi(\B{x}) := \operatorname{Ave} \left( \begin{bmatrix}
S(\B{x})\\
\operatorname{diag}\left(S(\B{x})S(\B{x})^\text{H}\right)
\end{bmatrix} \right).
\end{equation}
Non-Gaussian properties of $\B{x}$ can be detected through non-zero
coefficients of $\Phi$. Indeed, let us separate real coefficients and
potentially complex coefficients
$\Phi(\B{x})=\big(\Phi_{\text{real}}(\B{x}),
\Phi_{\text{complex}}(\B{x})\big)$, with $\Phi_{\text{real}}(\B{x})$
being the real coefficients $\operatorname{Ave}\big(|\B{W}\B{x}|,
|\B{W}\B{x}|^2, |\B{W}|\B{W}\B{x}||^2\big)$ and
$\Phi_{\text{complex}}(\B{x})$ being the remaining potentially complex
coefficients, that is the cross-layer correlations $\operatorname{Ave}
\left( \B{W}\B{x}\big(\B{W}|\B{W}\B{x}|\big)^\text{H} \right)$ or the second
layer correlations $\operatorname{Ave} \left(
\B{W}|\B{W}\B{x}|\big(\B{W}|\B{W}\B{x}|\big)^\text{H} \right)$ with
different scale correlation on the first wavelet operator.
\begin{proposition}
    If $\B{x}$ is Gaussian then $\Phi_{\text{complex}}(\B{x})\approx0$.
    \\
    If $\B{x}$ is time-symmetric, then
    $\mathrm{Im}\,\Phi_{\text{complex}}(\B{x})\approx0$.
\end{proposition}
More precisely, beyond detecting non-Gaussianity through non-zero
coefficients up to estimation error, $\Phi(\B{x})$ is able to quantify
different non-Gaussian behaviors, which will be crucial for source
separation. Appendix \ref{AppendixScatcovDashboard} presents a dashboard
that visualizes $\Phi(\B{x})$ and can be used to interpret signal
non-Gaussian properties such as sparsity, intermittency, and
time-asymmetry.

The dimensionality of the wavelet scattering covariance representation
depends on the number of scales $J$ considered i.e. the number of
wavelet filters of $\B{W}$. In order for largest scale coefficients to
be well estimated, one should choose $J \ll \log_2(d)$ where $d$ is
input data dimension. The maximum number of coefficients in $\Phi$ is
smaller than $\log^3_2(d)$ for $d\geq3$~\cite{morel2022scale}. Contrary
to higher dimensional representations or higher order statistics,
scattering covariance $\Phi(\B{x})$ are low-dimensional, low-order
statistics that can be efficiently estimated on a single realization of
a source and does not require tremendous amount of data for estimation
to converge~\cite{morel2022scale}. In other word, $\Phi$ is a low-variance representation.
This point is key for our source separation algorithm to be applied on
limited data. Wavelet scattering covariance $\Phi$ extracts average and
correlation features from a two-layer convolutional neural network with predefined wavelet filters.
It is analogous to the features extracted in \citet{gatys2015texture}
for generation, that considers however a pretrained convolutional neural
network. In the following we will also make use of the scattering
cross-covariance representation $\Phi(\B{x},\B{y})=\operatorname{Ave}
\operatorname{diag}\left(S(\B{x})S(\B{y})^\text{H}\right)$ that captures
scale dependencies across two signals $\B{x}$ and $\B{y}$.
\begin{proposition}
If $\B{x}$ and $\B{y}$ are independent then
\[
\Phi(\B{x},\B{y}) \approx 0
\]
\end{proposition}
The above proposition shows that $\Phi(\B{x},\B{y})$ detects independence up to estimation error, which will be useful when it comes to separating independent sources.

\section{Unsupervised Source Separation}
\label{source-separation}

To enable high-fidelity source separation in domains in which access to
training data---supervised or unsupervised---is limited, we cast source
separation as an optimization problem in a suitable feature space. Owing
to wavelet scattering covariance representation's ability to capture
non-Gaussian properties of multiscale stochastic processes without any
training, we perform source separation by solving an optimization
problem over the unknown sources using loss functions over wavelet
scattering covariance representations. Due to the inductive bias
embedded in the design of this representation space, we gain access to
interpretable features, which could further inform us regarding the
quality of the source separation process.

\subsection{Problem Setup}

Consider a linear mixing of unknown sources $\B{s}^{\ast}_i(t), \ i=1,\ldots,N$
via a mixing operator $\B{A}$,
\begin{equation}
\B{x}(t) = \B{A}\B{s}^{\ast}(t) + \Bs{\nu}(t) = \B{a}_1^\top \B{s}_1^{\ast}(t) + \B{n}(t),
\label{forward_model}
\end{equation}
with
\begin{equation}
\begin{aligned}
\B{s}^{\ast}(t) & = \left [\B{s}^{\ast}_1(t), \ldots, \B{s}^{\ast}_N(t)\right]^\top, \ \B{A} = \begin{bmatrix} \B{a}_1^\top \ \cdots \ \B{a}_N^\top
\end{bmatrix}, \\
\B{n}(t) & = \Bs{\nu}(t) + \sum_{i=2}^{N} \B{a}_i^\top \B{s}^{\ast}_i(t).
\end{aligned}
\label{forward_model_components}
\end{equation}
In the above expressions, $\B{x}(t)$ represents the observed data, and
$\Bs{\nu}(t)$ is the measurement noise. Here we capture the noise and
the mixture of all the sources except for $\B{s}^{\ast}_1(t)$ through
the mixing operator in $\B{n}(t)$ that does not longer depends on
$\B{s}^{\ast}_1(t)$. The matrices $\mathbf{x}(t)$ and $\mathbf{s}(t)$
have dimensions of $M \times T$ and $N \times T$, respectively,
where $T$ represents the number of time samples. The mixing operator
$\mathbf{A}$ has dimensions of $M \times N$. As a result, the product of
$\mathbf{a}_1^\top$ and $\mathbf{s}_1(t)$ yields a matrix of dimensions $M
\times T$, which corresponds to the contributions of source $\mathbf{s}_1(t)$
exclusively in $\mathbf{x}(t)$.

\textbf{Objective.} The aim is to obtain a point estimate $\B{s}_1(t)$
given a single observation $\B{x}(t)$ with the assumption that $\B{a}_1$
is known and that we have access to a few realizations
$\{\B{n}_k(t)\}_{k=1}^K$ as a training dataset. For example, in the case
of separating glitches from seismic data recorded during the NASA InSight mission, we will
consider $\B{n}_k(t)$ to be snippets of glitch-free data and $\B{a}_1$
to encodes information regarding polarization. We will drop the time
dependence of the quantities in equations~\eqref{forward_model}
and~\eqref{forward_model_components} for convenience.

\subsection{Principle of the Method}

The inverse problem of estimating $\B{s}_1$ from the given observed data
$\B{x}$, as presented in equation~\eqref{forward_model}, is ill-posed
since the solution is not unique. To constrain the solution space of the
problem, we incorporate prior knowledge in the form of realizations
$\{\B{n}_k\}_{k=1}^K$. We achieve this through a loss function that
emphasizes the wavelet scattering covariance representation of $\B{x} -
\B{a}_1^\top\B{s}_1$ to be close to that of $\B{n}_k,\ k=1,\ldots, K$:
\begin{equation}
\mathcal{L}_{\text{prior}} \left(\B{s}_1\right) := \frac{1}{K}\sum_{k=1}^K
\Big\|\Phi\big(\B{x} - \B{a}_1^\top\B{s}_1\big) - \Phi\big(\B{n}_k\big)\Big\|_2^2.
\label{scat_matching_prior}
\end{equation}
In the above expression, $\Phi$ is the wavelet scattering covariance
mapping as described in equation~\eqref{ScatteringCovariance}. With the prior loss defined, we impose data-consistency via:
\begin{equation}
\mathcal{L}_{\text{data}} \left(\B{s}_1\right) :=
\frac{1}{K}\sum_{k=1}^K
\Big\|\Phi\big(\B{a}_1^\top\B{s}_1
+ \B{n}_k\big)
- \Phi\big(\B{x}\big)\Big\|_2^2.
\label{scat_matching_data_consistency}
\end{equation}
The data consistency loss function $\mathcal{L}_{\text{data}}$ promotes
estimations of $\B{s}_1$ that for any training example from
$\{\B{n}_k\}_{k=1}^K$ the wavelet scattering covariance representation
of $\B{a}_1^\top\B{s}_1 + \B{n}_k$ is close to that of the observed
data.

To promote the independence of sources, we penalize the scattering cross-covariance between $\B{a}_1^\top\B{s}_1$ and $\B{n}_k$.
%
\begin{equation}
    \label{loss:indep}
    \mathcal{L}_{\text{cross}} (\B{s}_1) :=
    \frac{1}{K}\sum_{k=1}^K\Big\|\Phi\big(\B{a}_1^\top\B{s}_1, \B{n}_k\big)\Big\|_2^2,
\end{equation}
where $\Phi(\cdot, \cdot)$ is the scattering cross-covariance
representation (see section~\ref{scatcov_formulation}).

\subsection{Loss Normalization}

The losses described previously do not contain any weighting term for
the different coefficients of the scattering covariance representation.
We introduce in this section a generic normalization scheme, based on
the estimated variance of certain scattering covariance distributions.
This normalization, which has been introduced in \citet{refId0}, allows
to interpret the different loss terms in a standard form, and to include
them additively in the total loss term without overall loss weights. Let
us consider first the loss term given by
equation~\eqref{scat_matching_prior}, which compares the distance
between $\B{x} - \B{a}_1^\top\B{s}_1$ and available training samples
$\{\B{n}_k\}_{k=1}^K$ in the wavelet scattering representation space.
Specifying explicitly the sum on the $M$ wavelet scattering covariance
coefficients $\Phi_m, \ m=1, \ldots, M$, it yields
\begin{equation*}
\mathcal{L}_{\text{prior}}
\left(\B{s}_1\right) = \frac{1}{MK}
\sum_{m=1}^{M} \sum_{k=1}^K \Big|\Phi_m\big(\B{x} - \B{a}_1^\top\B{s}_1\big) - \Phi_m\big(\B{n}_k\big)\Big|^2.
\end{equation*}
Let us consider the second sum in this expression. In the limit where
$\Phi_m\big(\B{x} - \B{a}_1^\top\B{s}_1\big)$ is drawn from the same
distribution as $\{\Phi_m\big(\B{n}_k\big)\}_{k}^K$, the difference
$\Phi_m\big(\B{x} - \B{a}_1^\top\B{s}_1\big) - \Phi_m\big(\B{n}_k\big)$,
seen as a random variable, should have zero mean, and the same variance
as the distribution $\{\Phi_m\big(\B{n}_k\big)\}_{k}^K$ up to a factor
$2$. Denoting $\sigma^2\big(\Phi_m\big(\B{n}_k\big)\big)$ as this
variance, which can be estimated from
$\{\Phi_m\big(\B{n}_k\big)\}_{k}^K$, this gives a natural way of
normalizing the loss:
\begin{equation*}
\mathcal{L}_{\text{prior}} \left(\B{s}_1\right)  =
\frac{1}{MK}
\sum_{m=1}^{M} \sum_{k=1}^K \frac{\Big|\Phi_m\big(\B{x}
- \B{a}_1^\top\B{s}_1\big)
- \Phi_m\big(\B{n}_k\big)\Big|^2}{\sigma^2\big(\Phi_m\big(\B{n}_k\big)\big)}
\end{equation*}
or in a compressed form
\begin{equation}
    \label{loss_normalized_prior}
\mathcal{L}_{\text{prior}} \left(\B{s}_1\right) =
\frac{1}{K}\sum_{k=1}^K
\frac{ \Big\|\Phi\big(\B{x}
- \B{a}_1^\top\B{s}_1\big)
- \Phi\big(\B{n}_k\big)\Big\|_2^2}{\sigma^2\big(\Phi\big(\B{n}_k\big)\big)},
\end{equation}
which takes into account the expected standard deviation of each
coefficient of the scattering covariance representation. This
normalization allows for two things. First, it removes the normalization
inherent to the multiscale structure of $\Phi$. Indeed, coefficients
involving low  frequency wavelets tend to have a larger norm. Second, it
allows to interpret the loss value, which is expected to be at best of
order unity and to sum different loss terms of same magnitude.

We can introduce a similar normalization for the other loss terms. Loss
term (\ref{scat_matching_data_consistency}) should be normalized by the
$M$-dimensional vector $\sigma^2\big( \Phi\big(\B{a}_1^\top\B{s}_1 +
\B{n}_k\big) \big)$ that we approximate by
$\sigma^2\big(\Phi\big(\B{x}+\B{n}_k\big)\big)$, in order to have a
normalization independent on $\B{s}_1$, yielding
\begin{equation}
    \label{loss_normalized_data}
\mathcal{L}_{\text{data}} \left(\B{s}_1\right) :=
\frac{1}{K}\sum_{k=1}^K
\frac{\Big\|\Phi\big(\B{a}_1^\top\B{s}_1
+ \B{n}_k\big)
- \Phi\big(\B{x}\big)\Big\|_2^2}
{\sigma^2\big(\Phi\big(\B{x}+\B{n}_k\big)\big)} .
\end{equation}
Finally, loss term (\ref{loss:indep}) should be normalized by
$\sigma^2\big(\Phi\big(\B{a}_1^\top\B{s}_1, \B{n}_k\big)\big)$ that we
approximate by $\sigma^2\big(\Phi\big(\B{x},\B{n}_k\big)\big)$
\begin{equation}
    \label{loss_normalized_cross}
    \mathcal{L}_{\text{cross}} (\B{s}_1)
    =
    \frac{1}{K}\sum_{k=1}^K\frac{\Big\|\Phi\big(\B{a}_1^\top\B{s}_1, \B{n}_k\big)\Big\|_2^2}{\sigma^2\big(\Phi\big(\B{x},\B{n}_k\big)\big)},
\end{equation}
We can now sum the normalized loss terms defined in equations~\eqref{loss_normalized_prior}--\eqref{loss_normalized_cross}
to get the final optimization problem to perform source separation
\begin{equation}
\B{\widetilde{s}}_1 := \argmin_{\B{s}_1}\, \Big[
\mathcal{L}_{\text{data}} (\B{s}_1)
+ \mathcal{L}_{\text{prior}} (\B{s}_1)
+ \mathcal{L}_{\text{cross}} (\B{s}_1) \Big] .
\label{total_loss}
\end{equation}

Due to the delicate normalization of the three terms, we expect that
further weighting of the three losses using weighting hyperparameters is
not necessary. We propose to initialize the optimization problem in
equation~\eqref{total_loss} with $\B{s}_1 :=0$. Such choice means that
$\B{n}=\B{x}-a_1^\top\B{s}_1$ is initialized to $\B{x}$, which contains
crucial information on the sources, as will be explained in the next
section.

We have observed that as soon as we know the statistics $\Phi(\B{n})$,
our algorithm retrieves the unknown statistics of the source
$\Phi(\B{a}^{\top}_1\B{s}_1^{\ast})$. In other words the algorithm successfully
separates the sources in the scattering covariance space, this constitutes a convergence result, that can be proved under simplifying assumptions (see theorem \ref{main-theo}). Of course, in
many cases as we will see in the next section, our algorithm retrieves
point estimates of $\B{s}_1(t)$ that is stronger.
\begin{theorem}
Let $\mathbf{x}=\B{a}_1^\top \B{s}_1^{\ast} + \B{n}$ with $\mathbf{s}_1$ and $\mathbf{n}$ two independent processes.
Let us assume we have two processes $\widetilde{\mathbf{s}}_1$ and $\widetilde{\mathbf{n}}$ with $\mathbf{x}=\B{a}_1^\top \widetilde{\B{s}}_1 + \widetilde{\B{n}}$.

Under the following assumptions:
\begin{enumerate}[label=(\roman*)]
\label{main-theo}
    \item \label{assump:maxH_n} $\mathbf{n}$ has a maximum entropy distribution under moment constraints $\E\{\Phi(\mathbf{n})\}$
    \item \label{assump:maxH_nt} $\widetilde{\mathbf{n}}$ has a maximum entropy distribution under moment constraints $\E\{\Phi(\widetilde{\mathbf{n}})\}$
    \item\label{assump:samestat} $\E\{\Phi(\widetilde{\mathbf{n}})\}=\E\{\Phi(\mathbf{n})\}$
    \item\label{assump:indep} $\widetilde{\mathbf{s}}_1$ and $\widetilde{\mathbf{n}}$ are independent
    \item\label{assump:nonzeroFourier}
    The Fourier transform $\widehat{p}_\mathbf{n}$ of the distribution $p_\mathbf{n}$ of $\mathbf{n}$ is non-zero everywhere.
\end{enumerate}
one has $\mathbf{n}\overset{d}{=}\widetilde{\mathbf{n}}$ and $\B{a}_1^\top\mathbf{s}^{\ast}_1\overset{d}{=}\B{a}_1^\top\widetilde{\mathbf{s}}_1$ where the equality is on the distribution of the processes.
\end{theorem}
Essentially, it means that when the source $\B{n}$ is statistically
characterized by its scattering covariance descriptors the algorithm is
able to retrieve statistically the other sources. The theorem is proved and its assumptions are discussed in appendix~\ref{app:theorem}.
This
emphasizes the choice of a representation $\Phi$ that can approximate
efficiently the stochastic structure of multiscale processes
~\cite{morel2022scale}.

\section{Numerical Experiments}

The main goal of this paper is to derive a unsupervised approach to
source separation that is applicable in domain with limited access to
training data, thanks to the wavelet scattering covariance
representation. To provide a quantitative analysis to the performance of
our approach, we first consider a stylized synthetic example that
resembles challenges of real-world data. To illustrate how our method
performs in the wild, we apply our method to data recorded on Mars
during the InSight mission. We aim to separate transient thermally induced
microtilts, i.e., glitches~\cite{10.1029/2020EA001317,
barkaoui2021anatomy}, from the recorded data by the InSight lander's
seismometer. The code for partially reproducing the results can be found on \href{https://github.com/alisiahkoohi/insight_src_sep}{GitHub}. Our implementation is based on the \href{https://github.com/RudyMorel/scattering_covariance}{original PyTorch code} for wavelet scattering covariances~\cite{morel2022scale}.

\subsection{Stylized Example}

\begin{figure}[t]
    \centering

    \begin{subfigure}[b]{1.0\linewidth}
        \includegraphics[width=1.0\linewidth]{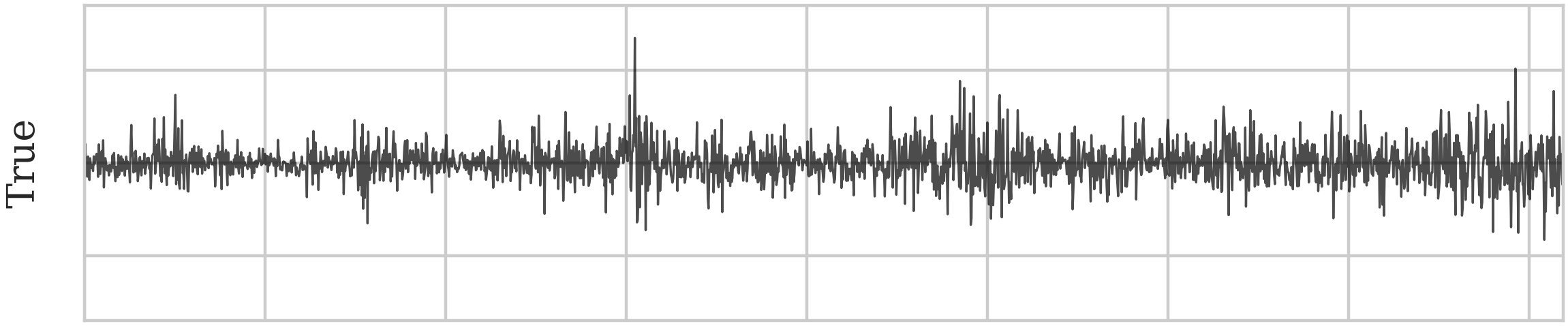}
    \end{subfigure}\hspace{0em}

    \begin{subfigure}[b]{1.0\linewidth}
        \includegraphics[width=1.0\linewidth]{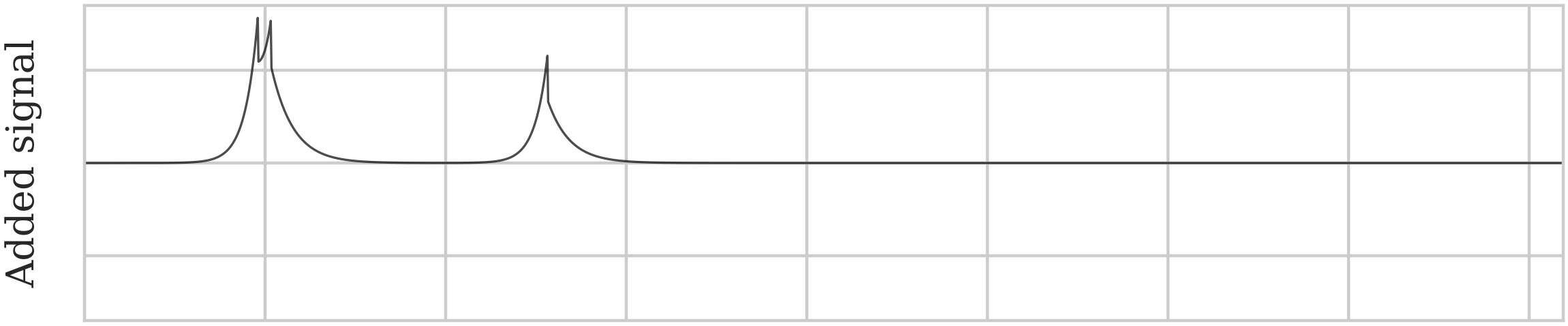}
    \end{subfigure}\hspace{0em}

    \begin{subfigure}[b]{1.0\linewidth}
        \includegraphics[width=1.0\linewidth]{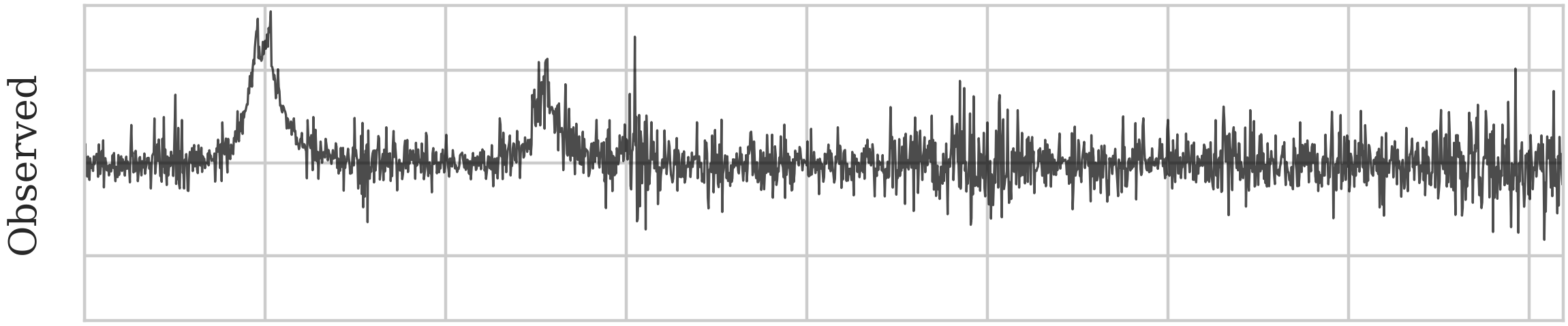}
    \end{subfigure}\hspace{0em}

    \begin{subfigure}[b]{1.0\linewidth}
        \includegraphics[width=1.0\linewidth]{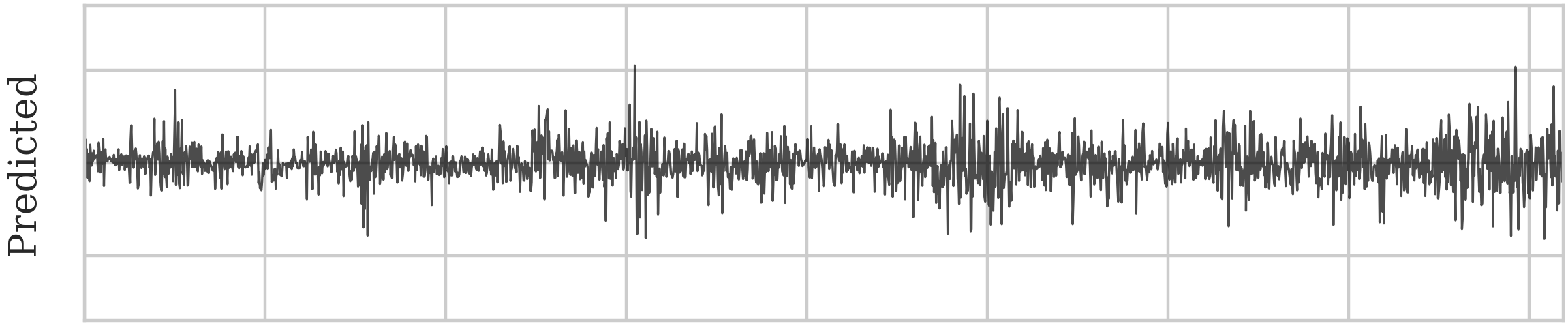}
    \end{subfigure}\hspace{0em}

    \begin{subfigure}[b]{1.0\linewidth}
        \includegraphics[width=1.0\linewidth]{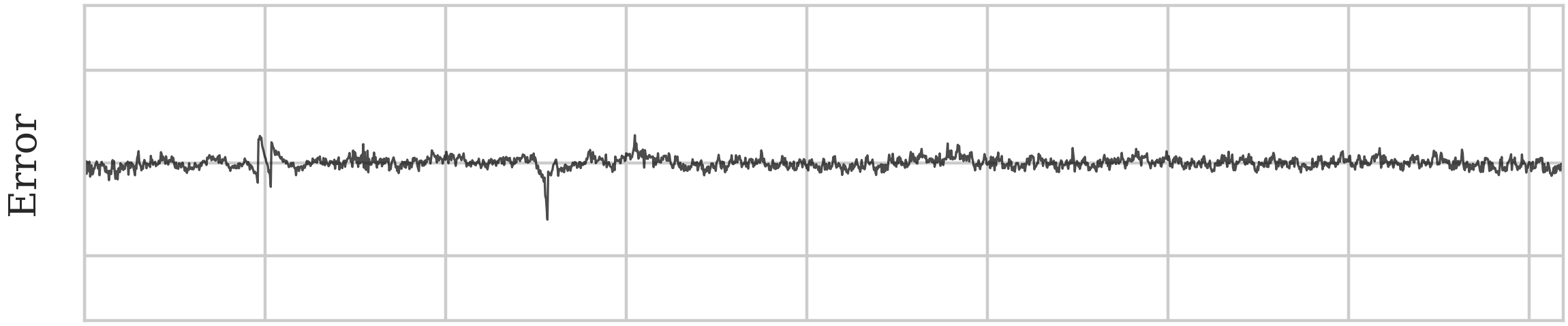}
    \end{subfigure}\hspace{0em}
    \caption{Unsupervised source separation applied to the multifractal
    random walk data. The vertical axis is the same for all the plots.}
    \label{fig:toy_example_1}
\end{figure}

We consider the problem of separating glitch-like signals from
increments of a multifractal random work
process~\cite{Bacry2001multifractal}. This process is a typical
non-Gaussian noise exhibiting long-range dependencies and showing bursts
of activity, e.g., see Figure~\ref{fig:synthetic_clean_samples} in the appendix for
several realizations of this process. The second source signal is
composed of several peaks with exponentially decaying amplitude, with
possibly different decay parameters on the left than on the right. To
obtain synthetic observed data, we sum increments of a multifractal
random walk realization, which plays the role of $\B{n}$ in
equation~\eqref{forward_model}, with a realization of the second source.
The top three images in Figure~\ref{fig:toy_example_1} are the signal of
interest, secondary added signal, and the observed data, respectively.

\begin{figure}
    \centering
    \subfloat{\includegraphics[width=1.0\linewidth]
    {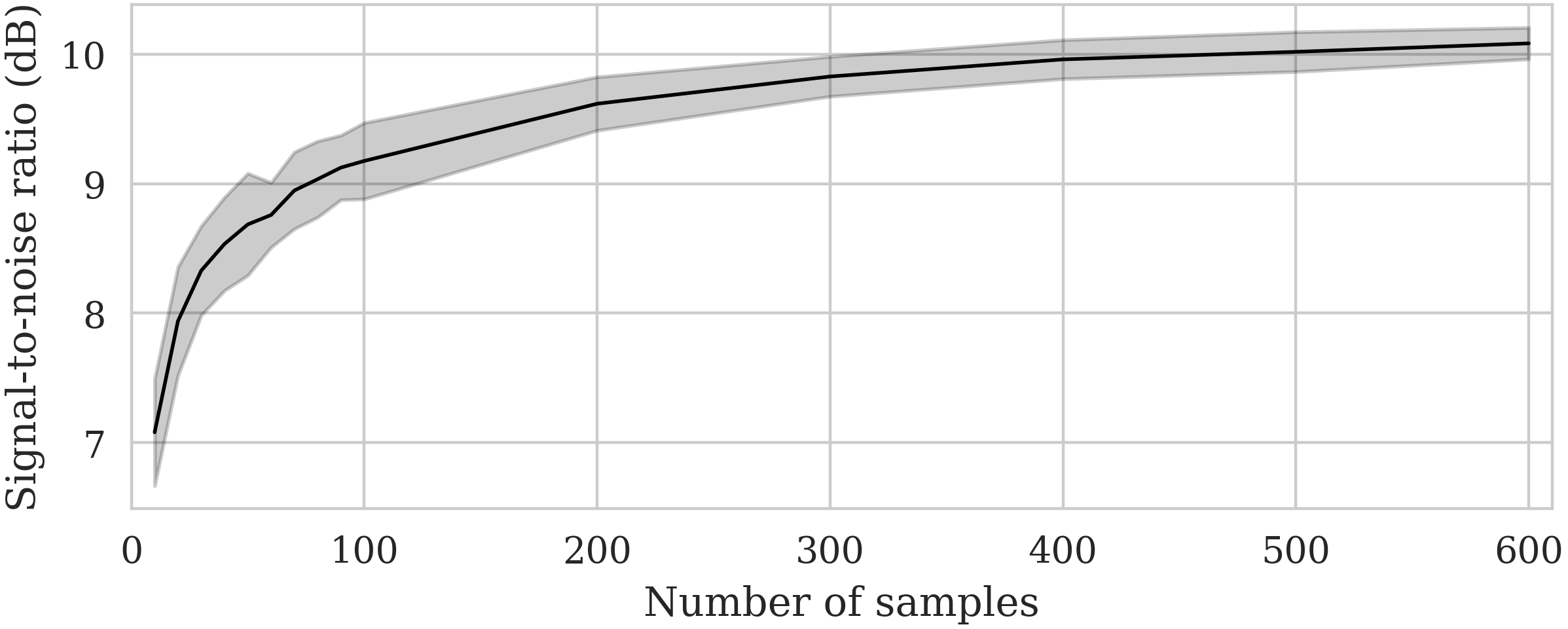}}
    \caption{Signal-to-noise ratio of the predicted multifractal random
    walk data versus number of unsupervised samples. Shaded area
    indicates the $90\%$ interval of this quantity for ten random source
    separation instances.}
    \label{fig:toy_SNR_vs_R}
\end{figure}

In order to retrieve the multifractal random walk realization, we solve
the optimization problem in equation~\eqref{total_loss} using the L-BFGS
optimization algorithm~\cite{Liu1989} using $500$ iterations. We use a
training dataset of $100$ realizations of increments of a multifractal
random walk, $\{\B{n}_k\}_{k=1}^{100}$. The architecture we use for
wavelet scattering covariance computation is two-layer scattering
network with $J=8$ different octaves with $Q=1$ wavelet per octave. We
use the same scattering network architecture throughout all the
numerical experiments in the paper. Given an input signal dimension of
$d=2048$, this choice of parameters yields a $174$-dimensional wavelet
scattering covariance space. The bottom two images in
Figure~\ref{fig:toy_example_1} summarizes the results. We are able to
recover the ground-truth multifractal random walk realization up to
small, mostly incoherent, and seemingly random error. To see the effect
of number of training realizations on the signal recovery, we repeated
the above examples and used varying number of training samples.
Figure~\ref{fig:toy_SNR_vs_R} shows that, as expected, the
signal-to-noise ratio of the recovered sources increases the more
training samples we have.

\begin{figure}[!t]
    \centering

    \begin{subfigure}[b]{1.0\linewidth}
        \includegraphics[width=1.0\linewidth]{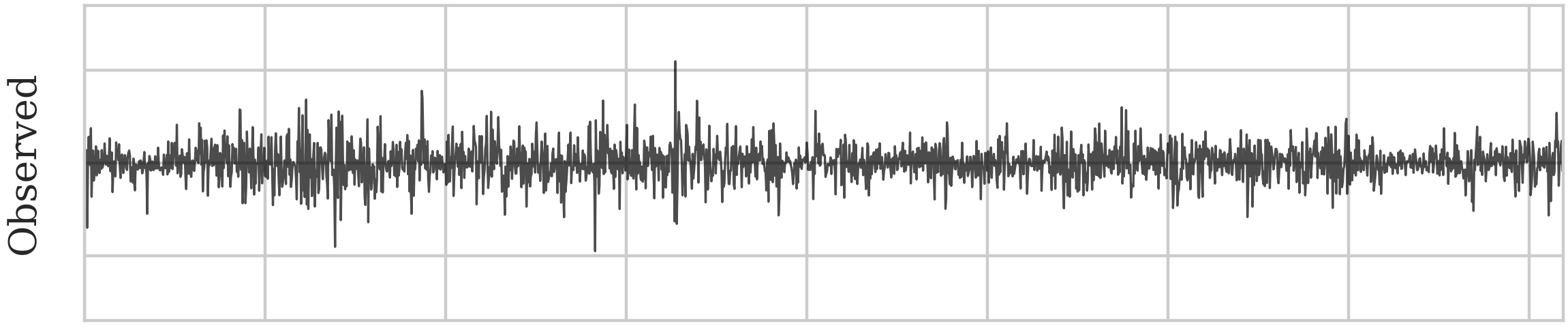}
    \end{subfigure}\hspace{0em}

    \begin{subfigure}[b]{1.0\linewidth}
        \includegraphics[width=1.0\linewidth]{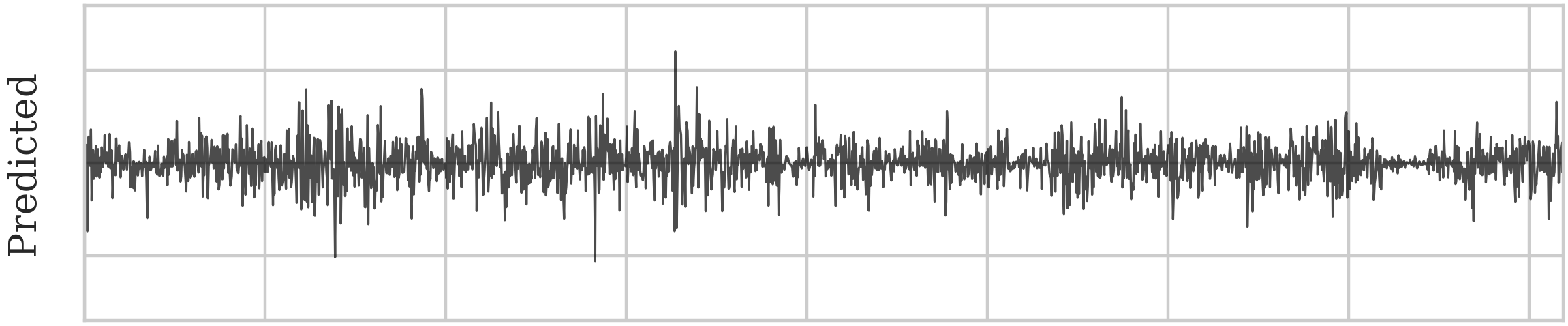}
    \end{subfigure}\hspace{0em}

    \begin{subfigure}[b]{1.0\linewidth}
        \includegraphics[width=1.0\linewidth]{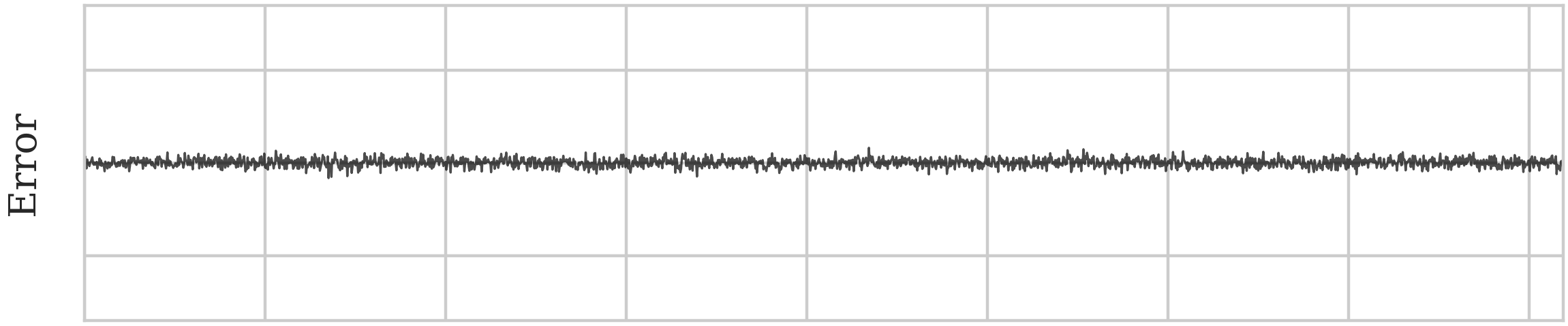}
    \end{subfigure}\hspace{0em}
    \caption{The behaviour when there are no sources to be removed, i.e., the observed data is a realization of the same stochastic process as the data snippets. The vertical axis is the same for all the plots.}
    \label{fig:corner_case}
\end{figure}

We also investigate the behaviour of our source separation algorithm in case there are no additional sources present in the signal, i.e., the observed data is a realization of the same stochastic process as the data snippets $\{\B{n}_k\}_{k=1}^K$. Ideally, the source separation algorithm should not unnecessarily remove important signals. We present the results of this experiment in Figure~\ref{fig:corner_case}, which indeed confirms that only a negligible amount of energy has been removed from the observed data in this case. We argue that the undesired separated signal from the observed data by our method is mainly due to errors in estimating the scattering covariance statistics using a finite amount of data snippets.

\begin{figure}[!t]
    \centering
    \subfloat{\includegraphics[width=1.0\linewidth]
    {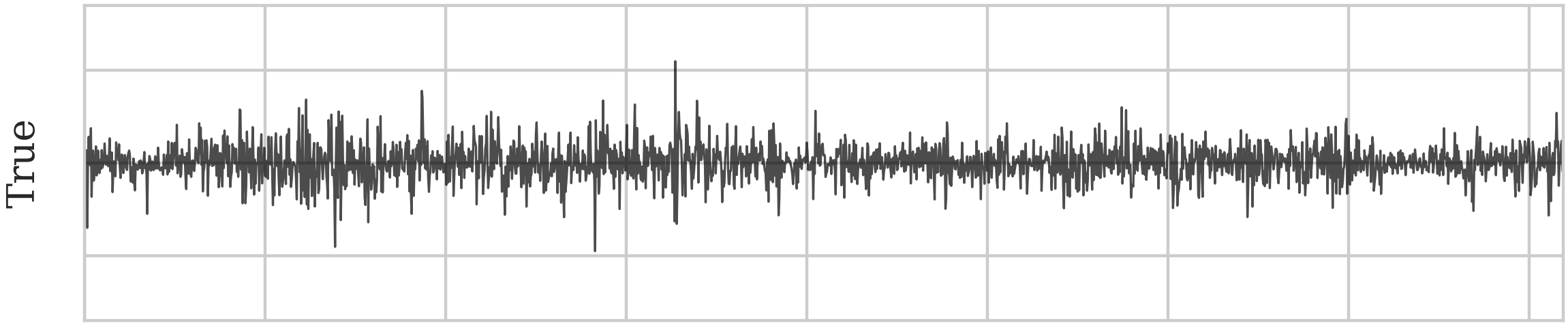}}
    \\
    \subfloat{\includegraphics[width=1.0\linewidth]
    {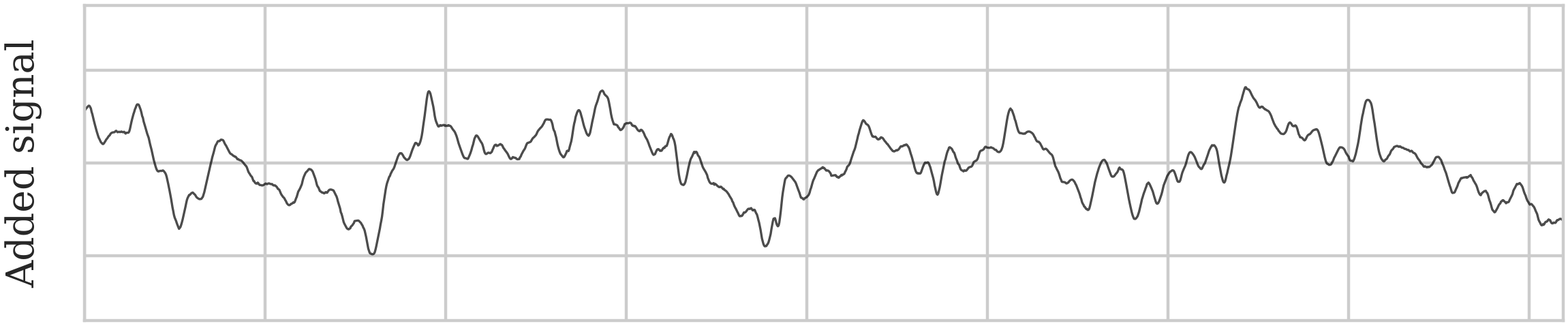}}
    \\
    \subfloat{\includegraphics[width=1.0\linewidth]
    {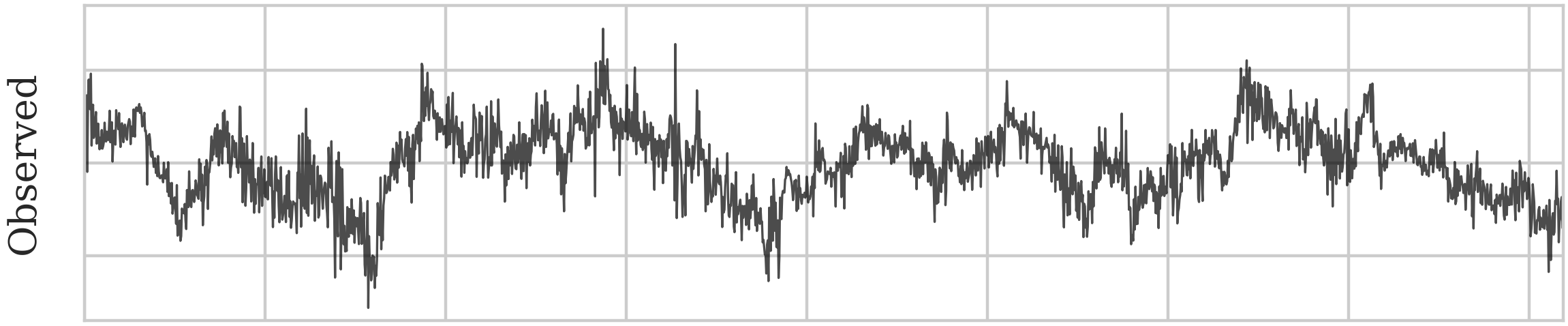}}
    \\
    \subfloat{\includegraphics[width=1.0\linewidth]
    {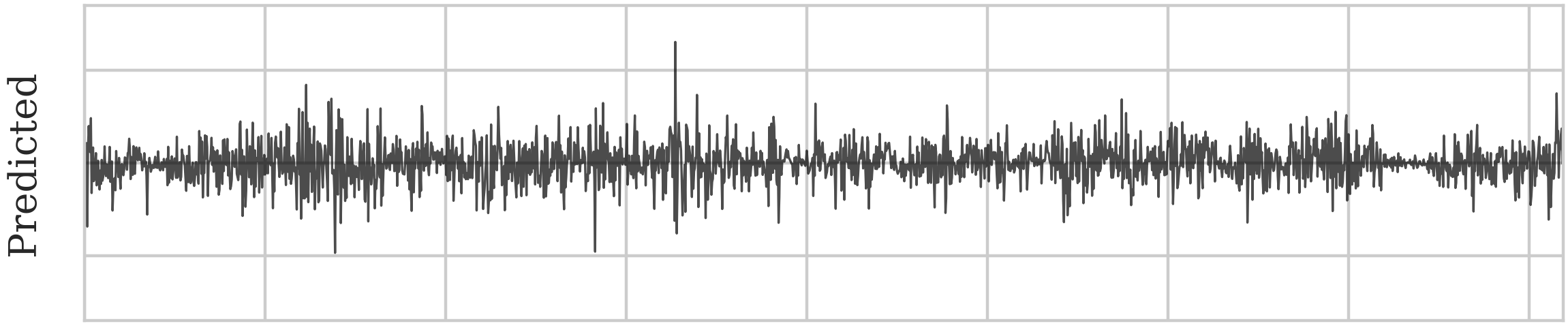}}
    \\
    \subfloat{\includegraphics[width=1.0\linewidth]
    {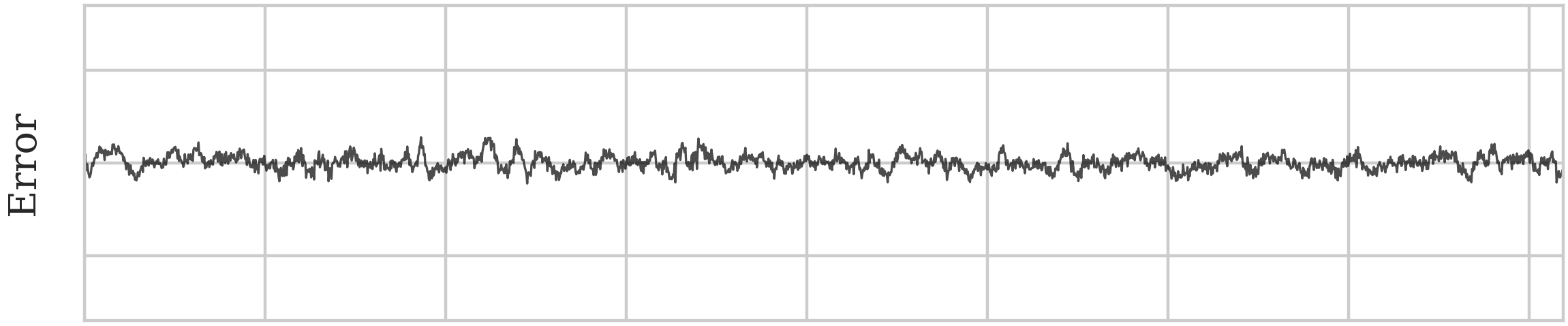}}
    \caption{Unsupervised source separation applied to the multifractal
    random walk data with a turbulent additive signal. The vertical axis
    is the same for all the plots.}
    \label{fig:toy_example_2}
\end{figure}

To show our method can also separate sources that are not localized in
time, we consider contaminating the multifractal random walk data with a
turbulent signal (see second image from the top in
Figure~\ref{fig:toy_example_2}. Without any prior knowledge regarding
this turbulent signal and by only using $100$ realizations of increments
of a multifractal random walk as training samples, we are able to
recover the signal of interest with arguably low error: juxtapose the
ground truth and predicted multifractal random walk realization in
Figure~\ref{fig:toy_example_2}. The algorithm correctly removes the low
frequencies content of the turbulent jet, and makes a small,
uncorrelated, random error at high frequencies. In this case the two
signals having different power spectra helps disentangling them at high
frequencies. In the above synthetic examples, the signal low frequencies
are well separated and the algorithm infers correctly the high
frequencies. In the earlier example, the presence of time localized
sources would facilitate the algorithm to "interpolate" the background
noise knowing its scattering covariance representation. This case makes
it more evident that the initialization $\B{s}_1=\B{0}$ informs the
algorithm of the trajectory of the unknown source.

\subsection{Application to Data from the InSight Mission}

InSight lander's seismometer, SEIS, is exposed to heavy wind and
temperature fluctuations. As a result, it is subject to background
noise. Glitches are a widely occurring family of noise caused by a
variety of causes~\cite{10.1029/2020EA001317}. These glitches often
appear as one-sided pulses in seismic data and significantly affect the
analysis of the data~\cite{10.1029/2020EA001317}. In this section we
will explore the application of our proposed method in separating
glitches and background noise from the recorded seismic data on Mars.

\subsubsection{Separating Glitches}
\label{sec:removing_glitches_main}

\begin{figure}[t]
    \centering
    \subfloat{\includegraphics[width=1.0\linewidth]
    {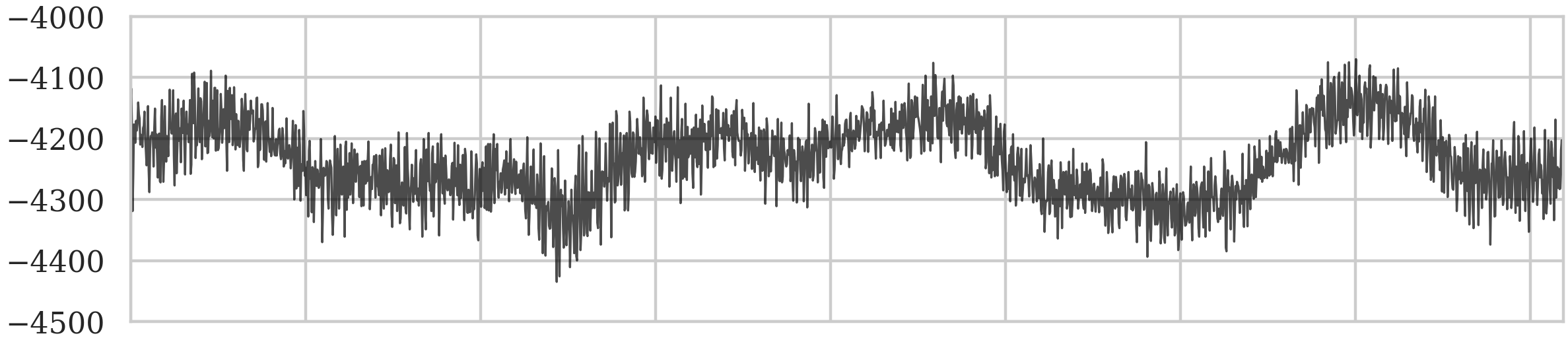}}
    \\
    \subfloat{\includegraphics[width=1.0\linewidth]
    {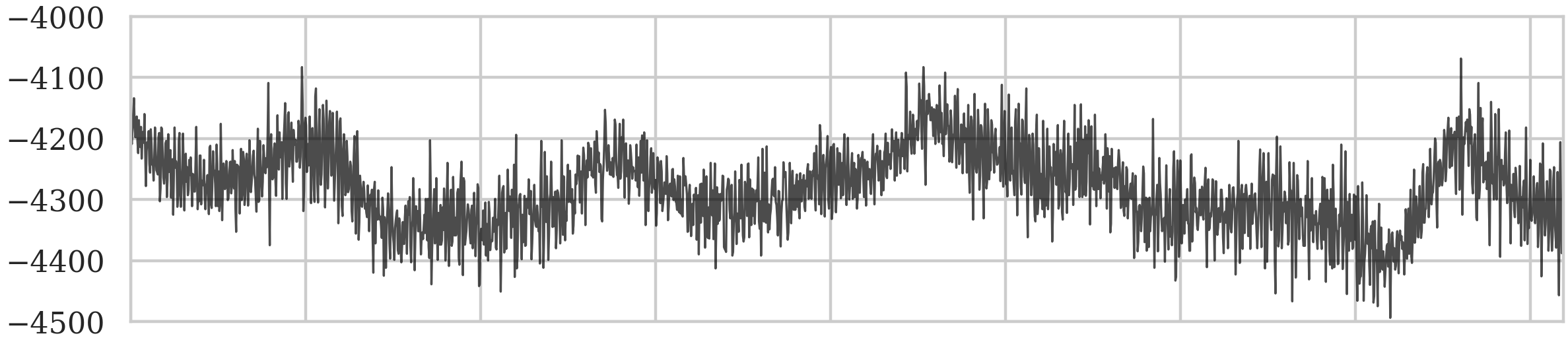}}
    \\
    \subfloat{\includegraphics[width=1.0\linewidth]
    {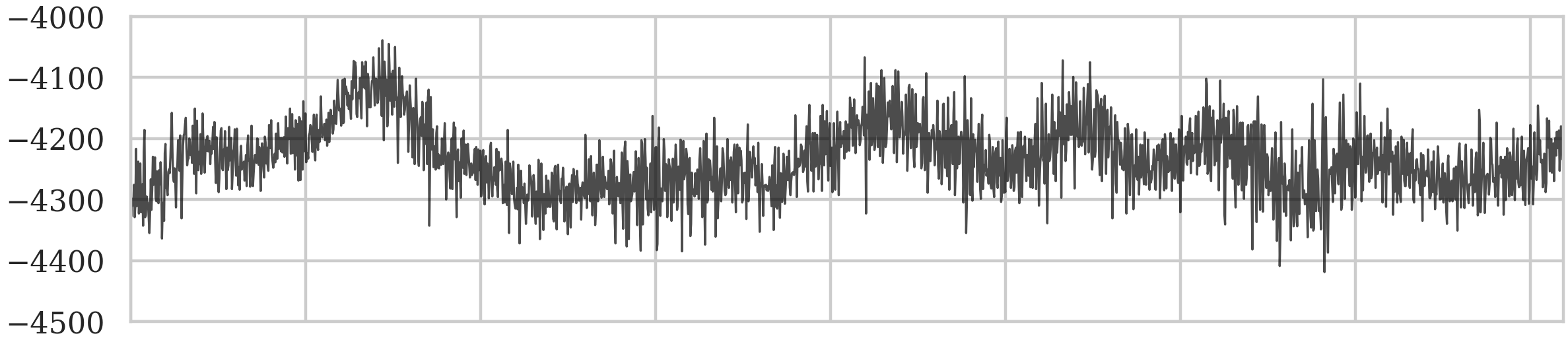}}
    \\
    \subfloat{\includegraphics[width=1.0\linewidth]
    {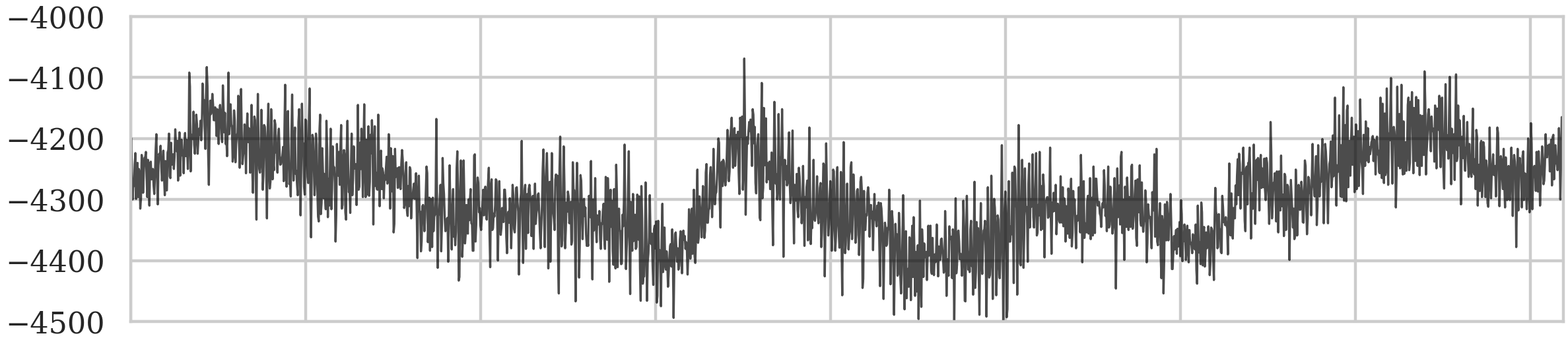}}
    \caption{Glitch-free snippets of the seismic data from Mars (U component).}
    \label{fig:mars_clean_samples}
\end{figure}

\begin{figure*}[t]
    \centering
    \subfloat{\includegraphics[width=0.495\linewidth]
    {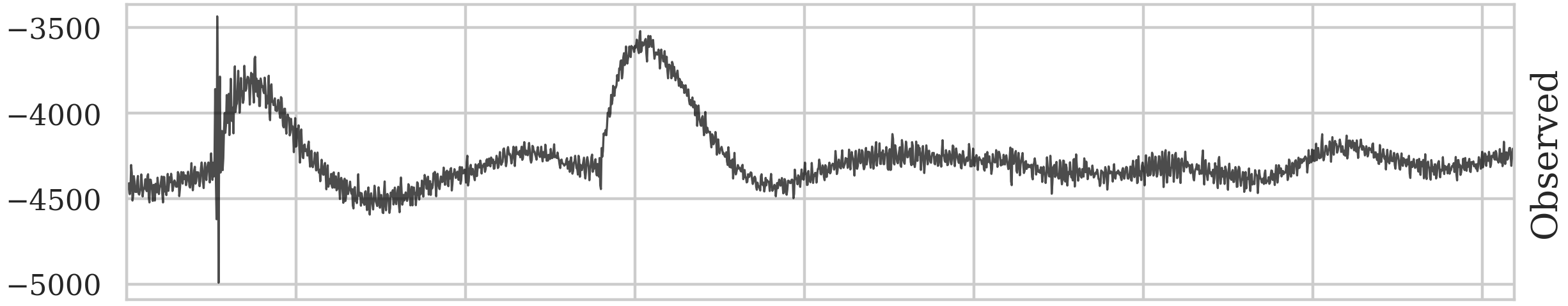}}
    \hspace{-0.00\linewidth}
    \subfloat{\includegraphics[width=0.495\linewidth]
    {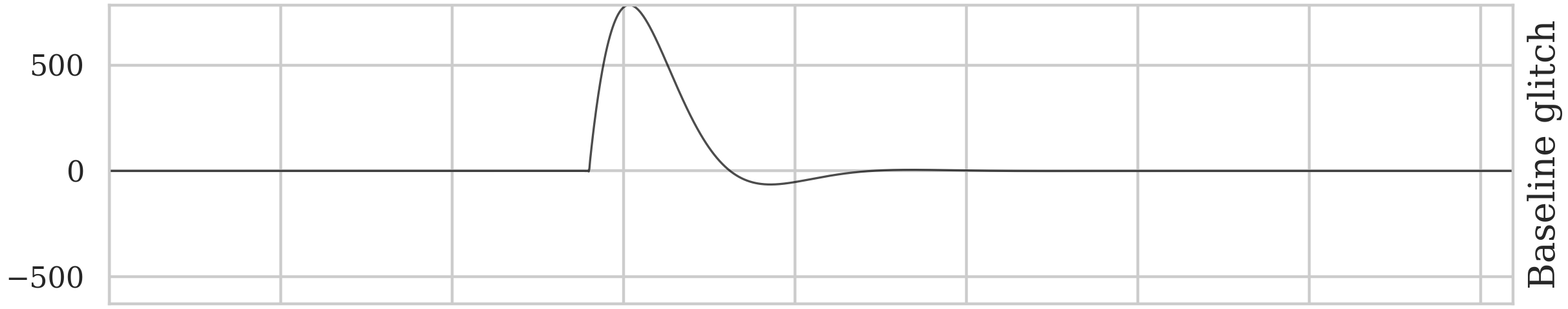}}
    \\
    \subfloat{\includegraphics[width=0.495\linewidth]
    {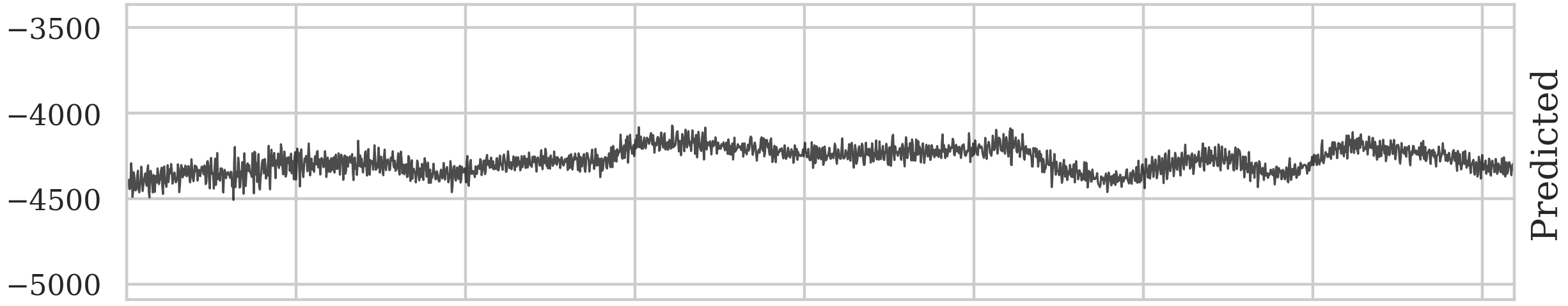}}
    \hspace{-0.00\linewidth}
    \subfloat{\includegraphics[width=0.495\linewidth]
    {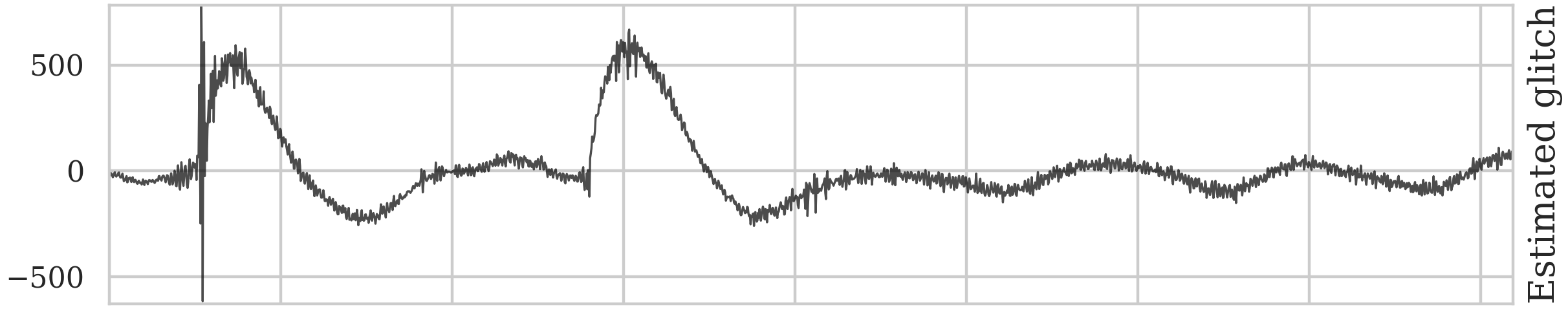}}
    \caption{Unsupervised source separation for glitch removal.
    Juxtapose the predicted glitches on the right. Our approach is able
    to remove a glitch whereas the baseline approach fails to detect
    it.}
    \label{fig:mars_example_1}
\end{figure*}

We propose to consider glitches as the source of interest $\B{s}_1$ in
the context of equation~\eqref{forward_model}. To perform source
separation using our technique, we need snippets of data that do not
contain glitches. We select these windows of data using an existing
catalog and glitches~\cite{10.1029/2020EA001317} and by further eye
examination to ensure no glitch contaminates our dataset. In total, we
collect 50 windows of length $102.4\,$s during sol 187 (6 June 2019) for
the U component. We show four of these windows of data in
Figure~\ref{fig:mars_clean_samples}. We perform optimization for glitch
removal using the same underlying scattering network architecture as the
previous example using 50 training samples and 1000 L-BFGS iterations.
Figure~\ref{fig:mars_example_1} summarizes the results. The top-left
image shows the raw data. Top-right image is the
baseline~\cite{10.1029/2020EA001317} (see Appendix~\ref{app:baseline} for description) prediction for the glitch signal.
Finally, the bottom row (from left to right) shows our predicted
deglitched data and the glitch signal separated by our approach. As
confirmed by experts at the InSight team, indeed our approach has
removed a glitch that the baseline has ignored (most likely due the
spike right at the beginning of the glitch signal). More
deglitching examples can be seen in
Figures~\ref{fig:mars_example_2}--\ref{fig:mars_example_5}.

It is important to note that the separated glitch in our experiments may comprise some
non-transient, non-seismic signals, potentially arising from atmosphere-surface
interactions, as opposed to the the baseline glitch. Consequently, we anticipate the separation of these non-seismic
signals in addition to the glitch when applying our approach. This results in ``noisy'' predicted glitches when compared to the baseline, which might be due to the the non-seismic signal. With this in mind, our approach extends the notion of glitch (as understood by the InSight
team). This is one of the benefits of our unsupervised approach as the method---based on the
statistics of the training data---identifies and removes events that do
not seem to belong to the training data distribution..

\begin{figure}[!t]
    \centering
    \includegraphics[width=0.9\linewidth]{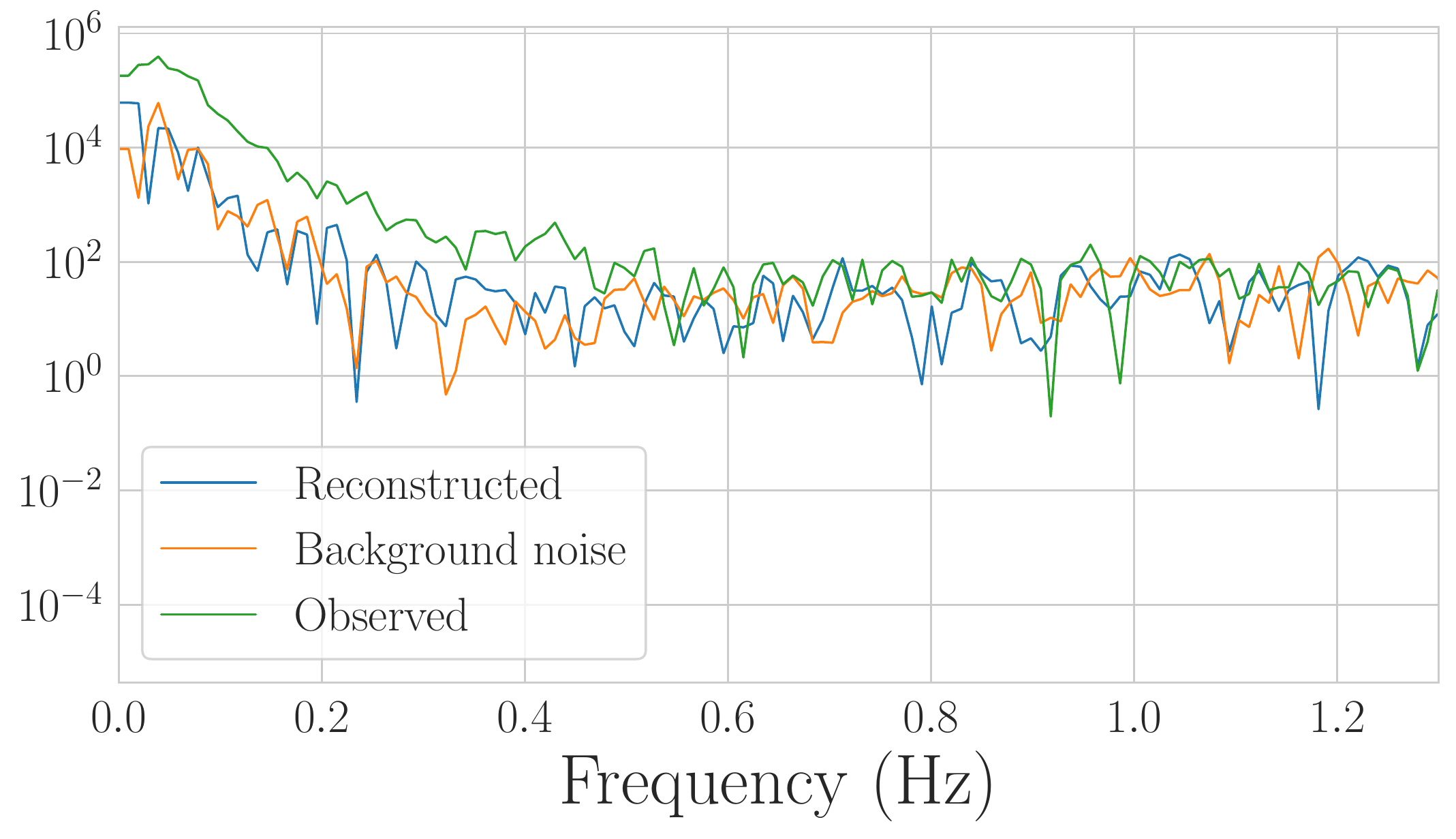}
    \caption{Power spectrum of the observed signal $\B{x}$, the
    background noise $\B{n}$ and the reconstructed background noise $\B{x}-a_1^\top\B{\widetilde{s}}_1$. We see that the reconstructed component
    statistically agrees with a Mars seismic background noise $\B{n}$.
    The algorithm efficiently removed the low-pass component of the
    signal corresponding to a glitch.}
    \label{fig:mars_power_spectrum}
\end{figure}

Thanks to the interpretability of wavelet scattering covariance
representations, stemming from our
comprehension of scattering coefficients and covariances, we can perform a source separation quality control in
domain where there is no access to ground truth source---as in our
example. Figure \ref{fig:mars_power_spectrum} compares the power spectra
of the reconstructed background noise (recorded data), a deglitched
realization of the background noise and the mixed signal (observed
data). It can be seen that the power spectrum of the background noise is
correctly retrieved. In fact, the scattering covariance statistics,
which extend the power spectrum, are correctly retrieved, which is due
to the loss term in equation~\eqref{scat_matching_prior}.

\subsubsection{Marsquake Background Noise Separation}

Marsquakes are of significant importance as they provide useful
information regarding the Mars subsurface, enabling the study of Mars'
interior~\cite{doi:10.1126/science.abf8966, doi:10.1126/science.abi7730,
doi:10.1126/science.abf2966}. Recordings by the InSight lander's
seismometer are susceptible to background noise and transient
atmospheric signals, and here we apply our proposed unsupervised source
separation approach to separate background noise from a
marsquake~\cite{doi.org/10.12686/a19}. To achieve this, we select about
30 hours of raw data (except for a detrending step)---from the U
component with a 20Hz sampling rate---to fully characterize various
aspects of the background noise through the wavelet scattering
covariance representation. Next, we window the data and use the windows
as training samples from background noise ($\B{n}_k$ in the context of
equation~\eqref{forward_model}) with the goal of retrieving the
marsquake recorded at February 3, 2022~\cite{doi.org/10.12686/a19}.

We use the same network architecture as previous examples to setup the
wavelet scattering covariance representation. We use a window size of
$204.8\,$s and solve the optimization problem in
equation~\eqref{total_loss} with $200$ L-BFGS iterations. The results
are depicted in Figure~\ref{fig:marsquake_cleaning}. There are clearly
two glitches that we have successfully separated, along with the
background noise. This results is obtained merely by using 30 hours of
raw data, allowing us to identify the marsquake as a separate source due
to differences in wavelet scattering covariance representation.

\section{Conclusions}

For source separation to be effective, prior knowledge concerning
unknown sources is necessary. Data-driven source separation methods
extract this information from existing datasets during pretraining. In
most cases, these methods require a large amount of data, which means
that they are not suitable for planetary science missions. To address
the challenge posed by limited data, we proposed an approach based on
wavelet scattering covariances. We reaped the benefits of the inductive
bias built into the scattering covariances, enabling us to obtain
low-dimensional data representations that characterize a wide range of
non-Gaussian properties of multiscale stochastic processes without
pretraining. Using a wavelet scattering covariance space optimization
problem, we were able to separate thermally induced microtilts
(glitches) from data recorded by the InSight lander's seismometer with
only a few glitch-free data samples. In addition, we applied the same
strategy to separate marsquakes from background noise and glitches using only several
hours of data with no recorded marsquake. Our approach did not
require any knowledge regarding glitches or marsquakes, and proved to be more
robust in separating glitches from recorded seismic data on Mars than existing
techniques. An important characteristic of our approach is that it serves as an exploratory method for unsupervised learning, particularly beneficial for investigating complex and real-world datasets.

\section{Acknowledgments}\label{acknowledgments}
\vspace*{-0.1cm}
Maarten V. de Hoop acknowledges support from the Simons Foundation under
the MATH + X program, the National Science Foundation under grant
DMS-2108175, and the corporate members of the Geo-Mathematical Imaging
Group at Rice University.

\newpage

\bibliography{example_paper}

\bibliographystyle{icml2023}

\newpage

\appendix

\section{Wavelet Scattering Covariance: Background Information}

\subsection{Wavelet Filters}
\label{AppendixWavelet}

A wavelet $\psi(t)$ has a fast decay away from $t=0$, polynomial or
exponential for example, and a zero-average $\int \psi(t)\,{\rm d}t =
0$. We normalize $\int |\psi(t)| \,dt = 1$. The wavelet transform
computes the variations of a signal $x$ at each dyadic scale $2^j$ with
\begin{equation*}
    \B{W}\B{x} (t,j) = \B{x}\star\psi_j(t)~~\mbox{where}~~\psi_j(t) = 2^{-j}\psi(2^{-j}t).
\end{equation*}
We use a complex wavelet $\psi$ having a Fourier transform
$\widehat{\psi}(\om) = \int \psi(t)\, e^{-i \om t}\, {\rm d}t$ which is
real, and whose energy is mostly concentrated at frequencies $\om \in
[{\pi}, {2}\pi]$. It results that $\widehat{\psi}_j(\omega) =
\widehat{\psi}(2^j\omega)$ is non-negligible mostly in $\om \in [2^{-j}
\pi, 2^{-j+1}\pi]$.

\begin{figure}[h]
    \centering
    \includegraphics[width=0.9\linewidth]{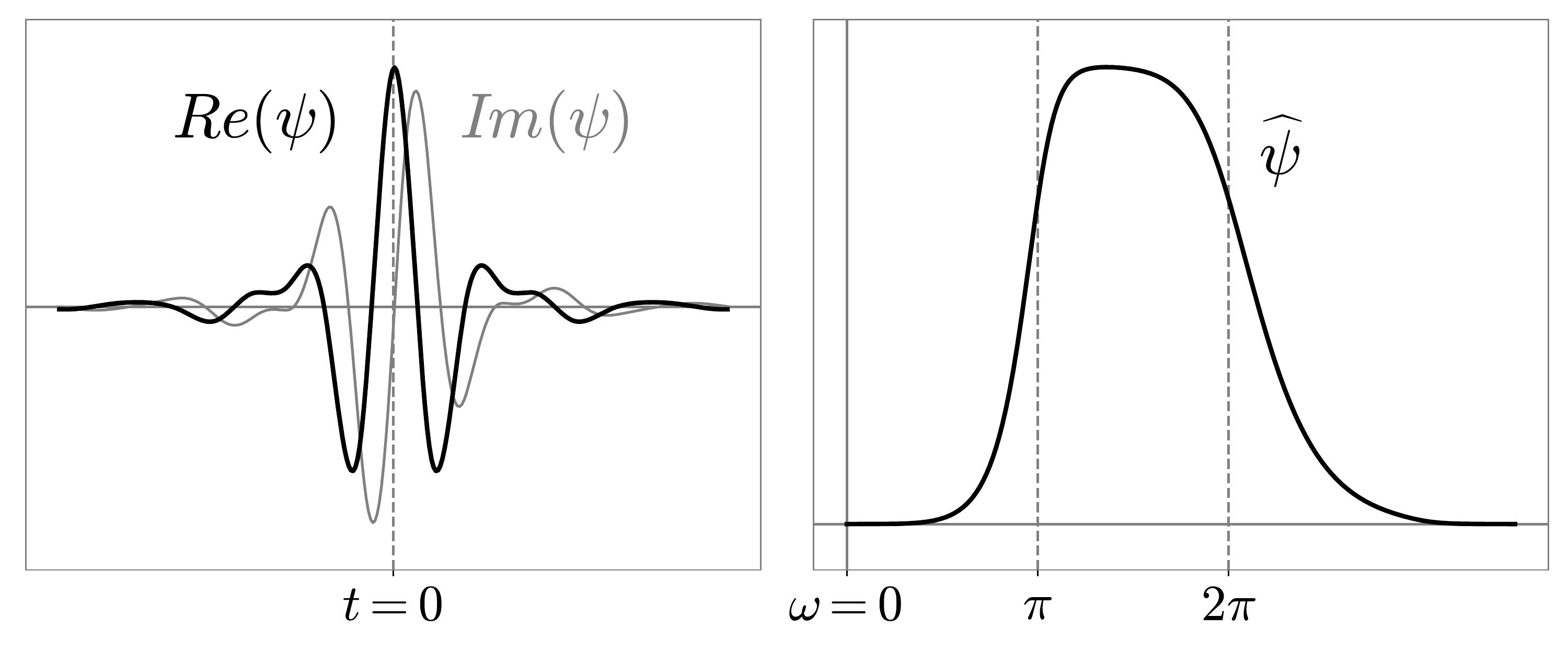}
    \caption{Left: complex Battle-Lemarié wavelet $\psi(t)$ as a
    function of $t$. Right: Fourier transform $\widehat{\psi}(\omega)$
    as a function of $\om$.}
    \label{fig:wavelet}
\end{figure}

We impose that the wavelet $\psi$ satisfies the following energy
conservation law called Littlewood-Paley equality
\begin{equation}
    \label{littlewood}
    \forall \omega > 0~~,~~ \sum_{j=-\infty}^{+\infty} |\widehat {\psi} (2^j \omega)|^2 = 1 .
\end{equation}
A Battle-Lemarié wavelet, see Figure \ref{fig:wavelet}, is an example of
such wavelet. The wavelet transform is computed up to a largest scale
$2^J$ which is smaller than the signal size $d$. The signal lower
frequencies in $[-2^{-J} \pi, 2^{-J} \pi]$ are captured by a low-pass
filter $\varphi_J(t)$ whose Fourier transform is
\begin{equation}
    \label{lowpass}
    \widehat{\varphi}_J (\omega) = {\Big(\sum_{j=J+1}^{+\infty} |\widehat{\psi} (2^j \omega)|^2\Big)^{1/2}} .
\end{equation}
One can verify that it has a unit integral $\int \varphi_J(t)\, {\rm d}t
= 1$. To simplify notations, we write this low-pass filter as a last
scale wavelet $\psi_{J+1} = \varphi_J$, and $\B{W}\B{x}(t,J+1) = \B{x}
\star \psi_{J+1}(t)$. By applying the Parseval formula, we derive from
(\ref{littlewood}) that for all $\B{x}$ with $\|\B{x}\|^2 = \int
|\B{x}(t)|^2\,dt < \infty$
\begin{equation*}
    \label{energyConservation}
    \|\B{W} \B{x} \|^2 = \sum_{j=-\infty}^{J+1}\|\B{x} \star \psi_j \|^2 = \|\B{x}\|^2 .
\end{equation*}
The wavelet transform $\B{W}$ preserves the norm and is therefore
invertible, with a stable inverse.

\subsection{Scattering Network Architecture}
\label{scatnet_arch}

A scattering network is a convolutional neural network with wavelet
filters. In this paper we choose a simple two-layer architecture with
modulus non-linearity:
\[
 S(\B{x}) := \begin{bmatrix}
\B{W}\B{x}\\
\B{W}|\B{W}\B{x}|
\end{bmatrix}.
\]
The wavelet operator $\B{W}$ is the same at the two layers, it uses
$J=8$ predefined Battle-Lemarié complex wavelets that are dilated from
the same mother wavelet by powers of $2$, yielding one wavelet per
octave.

The first layer extracts $J+1$ scale channels $\B{x}\star\psi_j(t)$, corresponding to $J$ band-pass and $1$ low-pass wavelet filters. The
second layer is $\B{W}|\B{W}\B{x}|(t;j_1,j_2) =
|\B{x}\star\psi_{j_1}|\star\psi_{j_2}(t)$. It is non-negligible only if
$j_1 < j_2$. Indeed, the Fourier transform of $|X \star \psi_{j_1}|$ is
mostly concentrated in $[-2^{-j_1} \pi, 2^{-j_1} \pi]$. If $j_2 \leq
j_1$ then it does not intersect the frequency interval $[2^{-j_2} \pi,
2^{-j_2+1} \pi]$ where the energy of $\widehat{\psi}_{j_2}$ is mostly
concentrated, in which case $S\B{x}(t;j_1,j_2) \approx 0$.

Instead of the modulus $|\cdot|$ we could use another non-linearity that
preserves the complex phase, however it does not improve significantly
the results in this paper.

\begin{figure*}[!h]
\centering
\includegraphics[width=0.75\linewidth]{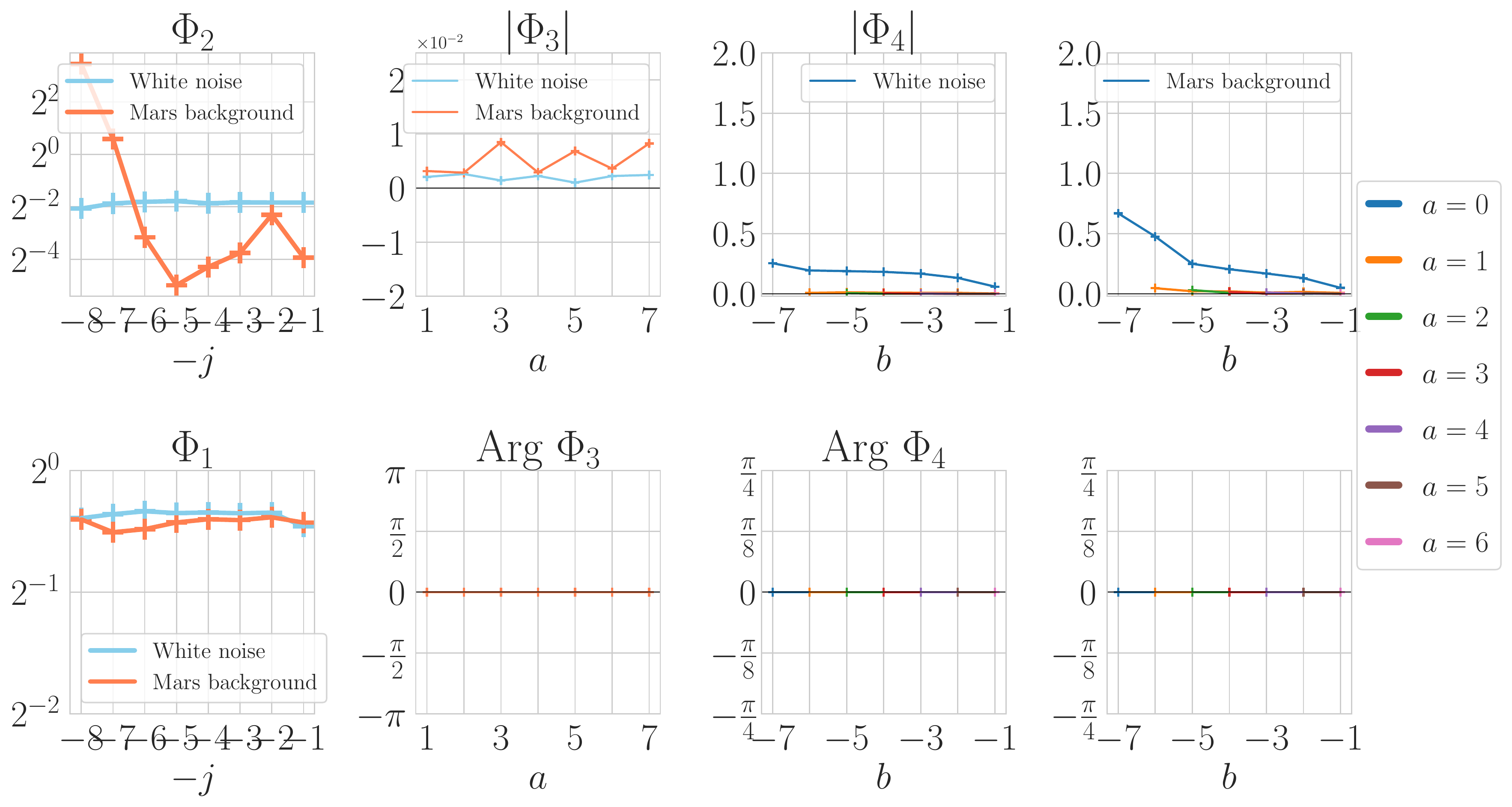}
\caption{Scattering covariance visualization of the Mars background
noise (no glitch) compared with a white noise. Estimation is
performed on the same amount of data.}
\label{fig:scattering_spectra_mars_simple}
\end{figure*}
\begin{figure*}[!h]
\centering
\includegraphics[width=0.75\linewidth]{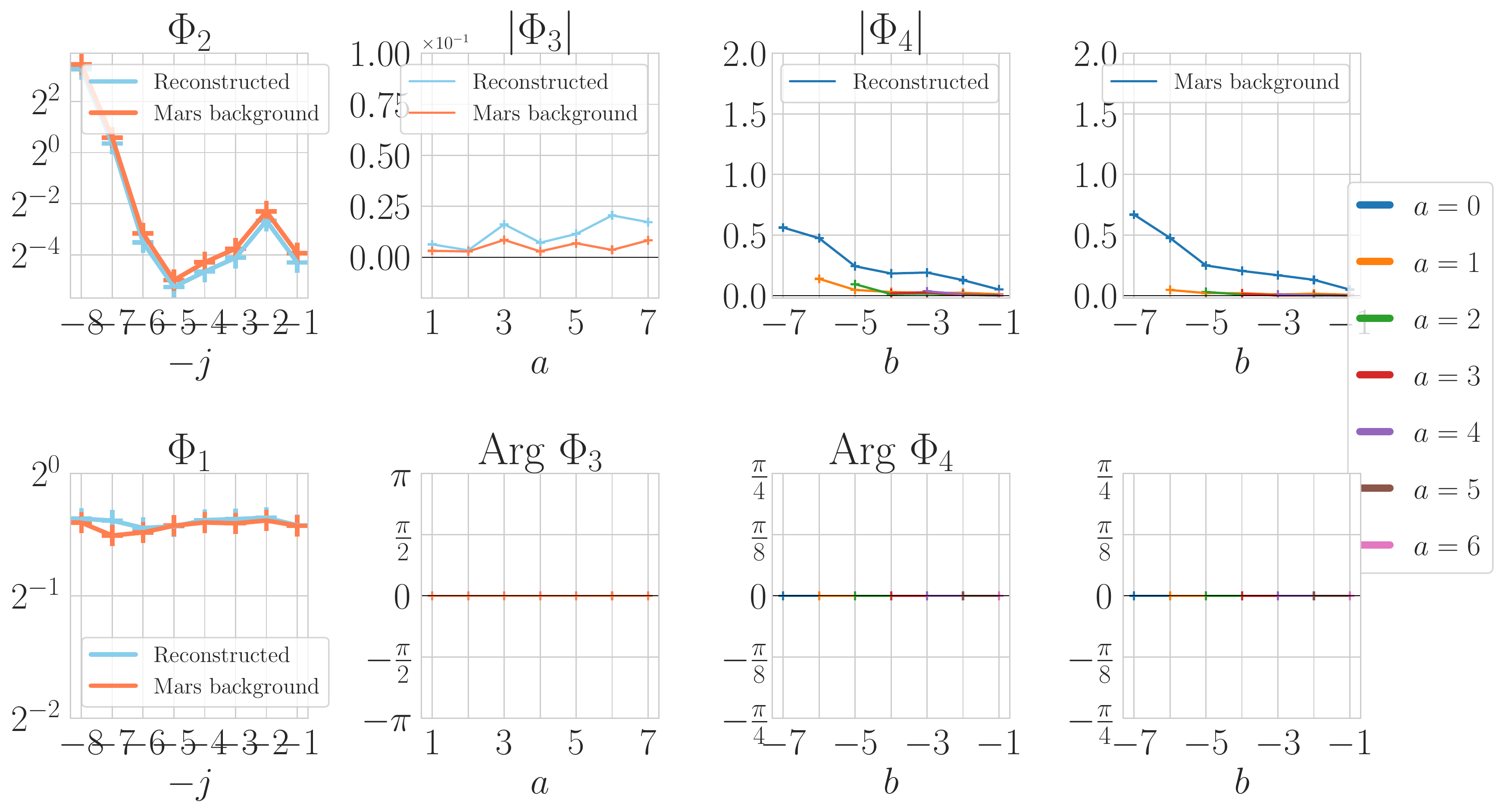}
\caption{Scattering covariance visualization of the reconstructed Mars background noise compared with a true Mars
background noise. This plots shows that beyond the wavelet power
spectrum, other non-Gaussian properties of the background noise such
as sparsity, long-range correlations match, up to a estimation
error.}
\label{fig:mars_scat_cov_match}
\end{figure*}

\subsection{Scattering Covariance Dashboard}
\label{AppendixScatcovDashboard}

The wavelet scattering covariance $\Phi(\B{x})$
(\ref{ScatteringCovariance}) contains four types of coefficients
$\Phi(\B{x}) = \big(\Phi_1(\B{x}), \Phi_2(\B{x}), \Phi_3(\B{x}),
\Phi_4(\B{x})\big)$. The first family provides $J$ order $1$ moment
estimators, corresponding to wavelet sparsity coefficients
\begin{equation}
    \label{sparsityFactor}
        \Phi_1(\B{x})[j] = \operatorname{Ave} |\B{x} \star \psi_j(t)|.
\end{equation}
The $J+1$ second order wavelet spectrum associated to $x$ are computed
by
\begin{equation}
    \label{waveletspectrum}
    \Phi_2(\B{x})[j] = \operatorname{Ave} \big(|\B{x} \star \psi_j(t)|^2\big).
    \end{equation}
There are $J(J+1)/2$ wavelet phase-modulus correlation coefficients for
$a>0$,
\begin{equation}
    \label{phaseEnvelopeCSpectrum}
        \Phi_3(\B{x})[j;a] = \operatorname{Ave} \big( \B{x} \star \psi_{j}(t)\,|\B{x} \star \psi_{j-a}(t)| \big).
\end{equation}
Finally, in total the scattering covariance includes $J(J+1)(J+2)/6$
scattering modulus coefficients for $a\geq0$ and $b<0$,
\begin{equation}
    \label{scatteringCSpectrum}
        \Phi_4(\B{x})[j;a,b] = \operatorname{Ave} \big(|\B{x}\star\psi_{j}|\star\psi_{j-b} (t)\, |\B{x}\star\psi_{j-a}|\star\psi_{j-b}^*(t)\big).
\end{equation}
These coefficients extend the standard wavelet power spectrum
$\Phi_2(\B{x})$. After appropriate normalization and reduction that we
describe below, scattering covariances can be visualized as a dashboard that displays non-Gaussian properties of $\B{x}$,
which is shown for example in Figures~\ref{fig:scattering_spectra_mars_simple} and~\ref{fig:mars_scat_cov_match}.

The power spectrum $\Phi_2(x)$ is plotted in a standard way, it is the
energy of the scale channels of $\B{x}\star\psi_j(t)$. This energy
affects the other coefficients
$\Phi_1(\B{x}),\Phi_3(\B{x}),\Phi_4(\B{x})$. To deduct this influence,
we normalize these coefficients by the power spectrum, $\Phi_1(\B{x})[j]
/ \sqrt{\Phi_2(\B{x})[j]}$, $\Phi_3(\B{x})[j;a] /
\sqrt{\Phi_2(\B{x})[j]\Phi_2(\B{x})[j-a]}$ and $\Phi_4(\B{x})[j;a,b] /
\sqrt{\Phi_2(\B{x})[j]\Phi_2(\B{x})[j-a]}$. Finally, we average
$\Phi_3(x)$ and $\Phi_4(x)$ on $j$, in order to plot scaling invariant
quantities, which reduces the number of coefficient to visualize~\cite{morel2022scale}.

\section{Source Separation Guarantees}
\label{app:theorem}

We prove theorem \ref{main-theo}, discuss its assumptions for the deglitching example applied to data from Mars, and show how our implementation relates to these assumptions. For sake of simplicity we take $a_1=1$.
\begin{proof}[Proof] Part I. One can prove that there exists a unique process $\mathbf{n}$ that maximises entropy under moment constraint $\E\{\Phi(\mathbf{n})\}$,
its distribution takes the form $p_\mathbf{n}(\cdot) = Z^{-1}_{\boldsymbol{\theta}}e^{-\boldsymbol{\theta}^\top \Phi(\cdot)}$ for certain Lagrange multipliers $\boldsymbol{\theta}\in{\mathbb R}^M$ where $M$ is the dimension of $\Phi$.
Assumptions \ref{assump:maxH_n}, \ref{assump:maxH_nt}, \ref{assump:samestat} imply that $\mathbf{n}$ and $\widetilde{\mathbf{n}}$ are the same unique process, meaning $p_\mathbf{n} = p_{\widetilde{\mathbf{n}}}$.
\\
Part II. Due to the independence of $\B{s}_1^{\ast},\mathbf{n}$ and $\widetilde{\mathbf{s}}_1,\widetilde{\B{n}}$ \ref{assump:indep} we have $p_{\textbf{x}}=p_{\B{s}_1^{\ast}}\star p_{\B{n}}$ and $p_{\textbf{x}}=p_{{\widetilde s}_1}\star p_{\widetilde{\B{n}}}$.
Since $p_{\widetilde{\mathbf{n}}}=p_\mathbf{n}$ we get $p_{\B{s}_1^{\ast}}\star p_\mathbf{n} = p_{\widetilde{\mathbf{s}}_1}\star p_{\widetilde{\mathbf{n}}}$.
This is a measure deconvolution problem.
Taking the Fourier transform on measures yields
\[
(\widehat{p}_{\B{s}_1^{\ast}} - \widehat{p}_{\widetilde{\mathbf{s}}_1}) \,  \widehat{p}_\mathbf{n}= 0.
\]
Under assumption \ref{assump:nonzeroFourier} we get $p_{\widetilde{\mathbf{s}}_1}=p_{\B{s}_1^{\ast}}$, which proves the theorem.
\end{proof}

Assumption \ref{assump:maxH_n} is the main assumption.
It implies that the processes $\mathbf{n}$ is fully determined by the values $\E\{\Phi(\mathbf{n})\}$, since there is a unique distribution satisfying \ref{assump:maxH_n}.
A maximum entropy process $\mathbf{n}$ under correlation constraints $\E\{\mathbf{n} \mathbf{n}^\top\}$ is a Gaussian process.
A wavelet Scattering Covariance captures non-linear correlations,
assumption \ref{assump:maxH_n} tells us that process $\mathbf{n}$ is a non-Gaussian noise fully characterized by $\E\{\Phi(\mathbf{n})\}$.
Now, the Scattering Covariance $\E\{\Phi(\mathbf{n})\}$ was shown to characterize a wide range of non-Gaussian noises (\cite{morel2022scale}).
In our case, the Mars seismic background noise $\mathbf{n}$ may not be fully characterized by its Scattering Covariance $\E\{\Phi(\mathbf{n})\}$, so that assumption \ref{assump:maxH_n} is only verified approximately, depending on the descriptive power of the representation $\E\{\Phi(\mathbf{n})\}$ for $\mathbf{n}$.

Assumption \ref{assump:maxH_nt} is approximately verified, requiring the entropy of $\mathbf{x}$ to be close to the entropy of $\mathbf{n}$, which is typically the case of time-localized signals such as glitch, of comparable amplitude than $\mathbf{n}$.
The gradient descent algorithm implements \ref{assump:maxH_nt}, reconstructed $\widetilde{\mathbf{n}}$ is initialized to $\mathbf{x}$ and is updated until $\Phi(\mathbf{x})$ matches the $\Phi(\mathbf{n}_k)$.

Assumption \ref{assump:samestat} is imposed through the loss term $\mathcal{L}_{\text{prior}}$, up to estimation error of $\Phi(\mathbf{n})$ on a finite number of realizations.

Assumption \ref{assump:indep} relates to the loss term $\mathcal{L}_{\text{cross}}$ that imposes statistical independence up to the cross-Scattering Covariance.

Assumption \ref{assump:nonzeroFourier} is a technical assumption satisfied for a Gaussian noise $\mathbf{n}$ for which the Fourier transform of $p_\mathbf{n}$ is a Gaussian.
A non-Gaussian noise $\mathbf{n}$ satisfying \ref{assump:maxH_n} has a distribution of the form $p_\mathbf{n}(\cdot) = Z^{-1}_{\boldsymbol{\theta}}e^{-\boldsymbol{\theta}^\top \Phi(\cdot)}$.
Apart from the coefficients $\operatorname{Ave}(S(\mathbf{n}))$, the scattering covariance $\Phi$ is quadratic in $\mathbf{n}$, thus we may assume \ref{assump:nonzeroFourier} is still satisfied.

\section{Baseline Method}
\label{app:baseline}

The glitch detection algorithm that we use as baseline is developed by \citet{10.1029/2020EA001317} and consists of several processing steps applied to seismic data:

\begin{itemize}
    \item Decimation: The data is downsampled to a uniform rate of two samples per second to ensure consistent parameterization and improve computational efficiency;

    \item Deconvolution and band-pass filtering: Instrument response is removed from each component, transforming the data into acceleration. Additional band-pass filtering is also applied to highlight the significant features of acceleration;

    \item Time derivative calculation: The time derivative of the filtered acceleration data is computed, resulting in acceleration steps becoming impulse-like signals;

    \item Glitch detection: A constant threshold is applied to the time derivative to identify glitches. A window length is introduced to avoid false triggers on subsequent samples that are part of the same glitch event, serving as a safeguard against spurious detections.
\end{itemize}

After glitch detection, removal is based on obtaining a model (template) for the glitch signatures, followed by a separation techniques that assumes the observed data as a linear combination of the glitch and the glitch spike. To characterize each detected glitch, a glitch model is employed, consisting of three parameters: an amplitude scaling factor, an offset, and a linear trend parameter. The modeling process entails solving a nonlinear least squares data fitting problem to determine these parameters. Subsequently, the deglitched data is obtained by subtracting the fitted glitch (excluding the offset and linear trend) from the original data.

In comparison to our approach, the glitch modeling step in the mentioned method could be a significant limitation. Unlike their method, we do not make any assumptions about the functional form of the glitch or the unknown source. Instead, we focus on learning the wavelet scattering covariance statistics of the background noise. This allows us to overcome the potential limitations associated with explicitly modeling the glitches.

\section{Multifractal Random Walk Realizations}
\label{random_walk_realizations}

\begin{figure*}[t!]
    \centering
    \subfloat{\includegraphics[width=0.5\linewidth]
    {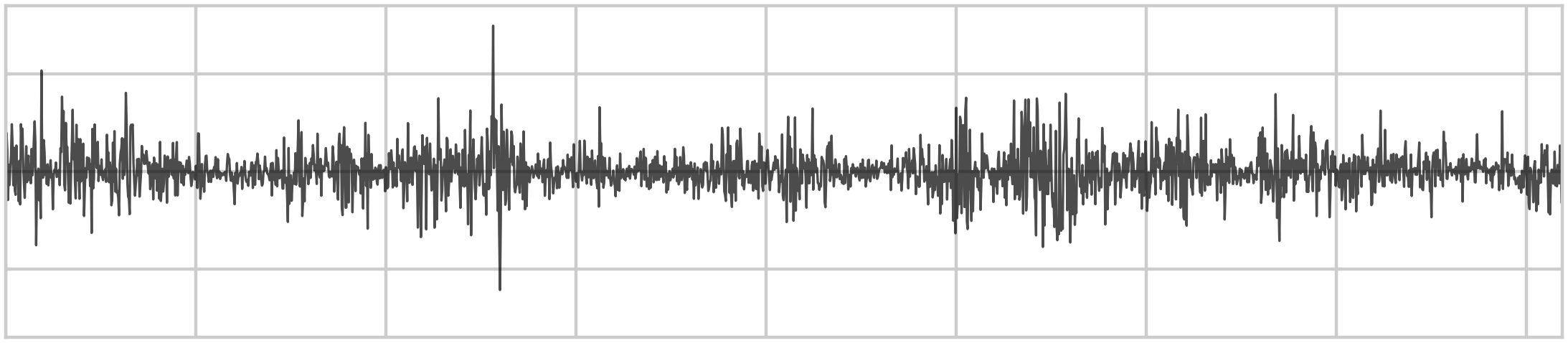}}
    \subfloat{\includegraphics[width=0.5\linewidth]
    {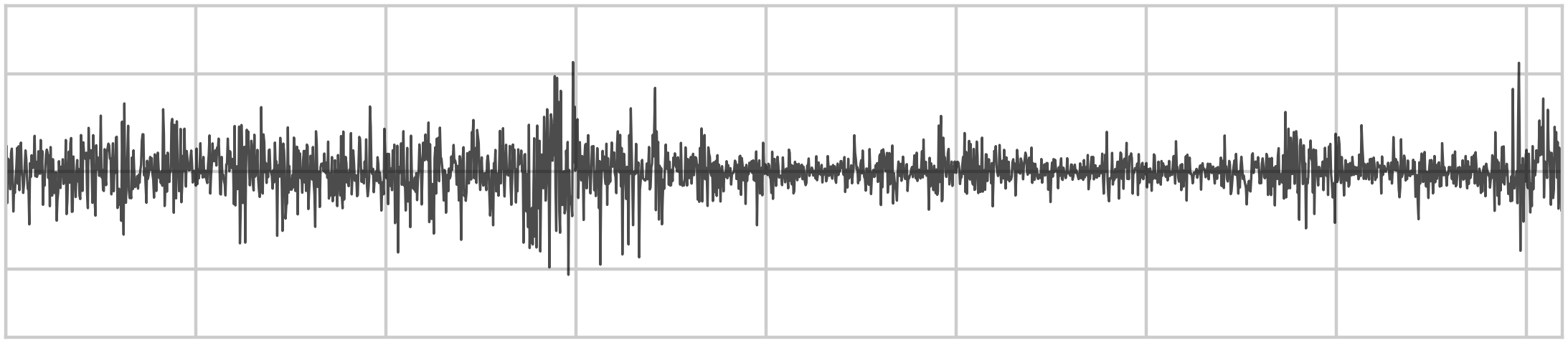}}
    \\
    \subfloat{\includegraphics[width=0.5\linewidth]
    {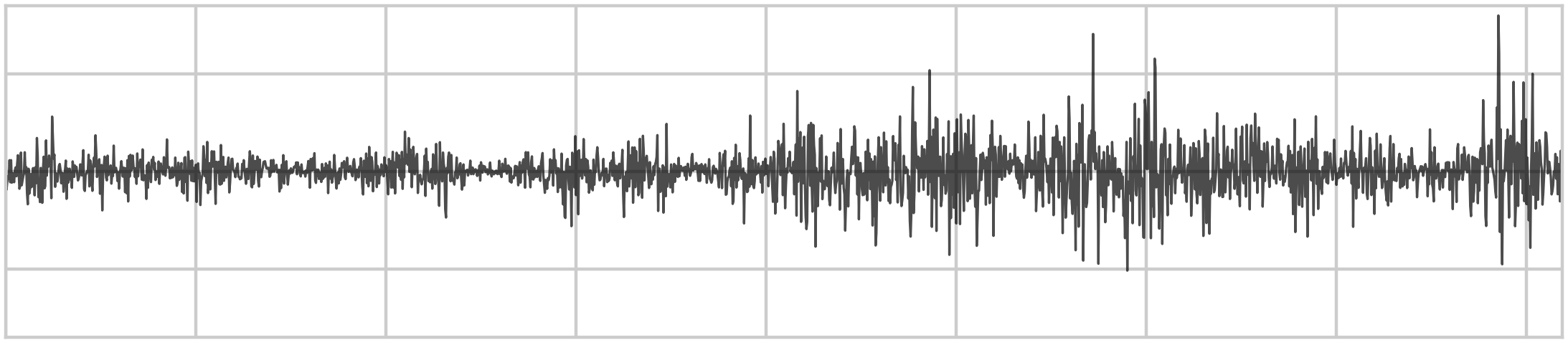}}
    \subfloat{\includegraphics[width=0.5\linewidth]
    {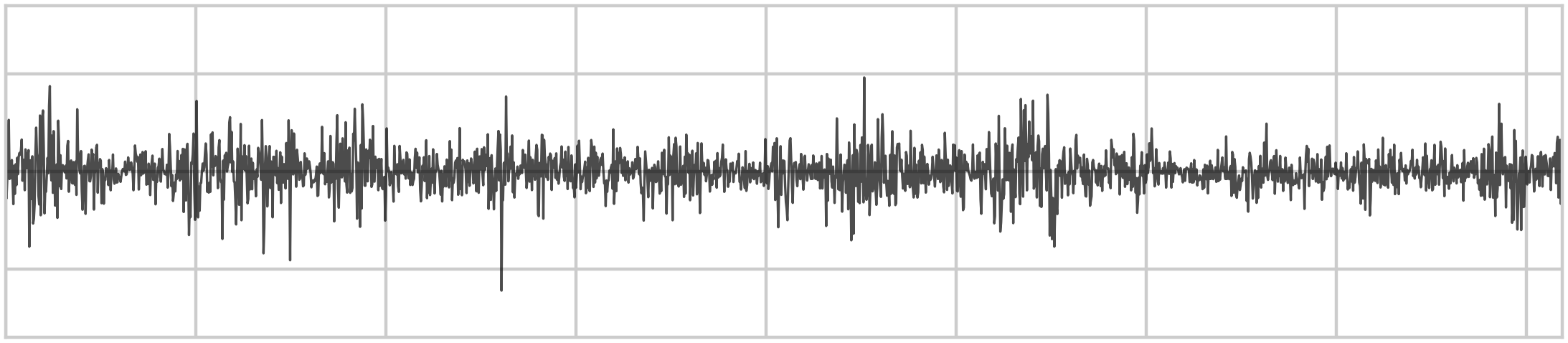}}
    \caption{Realizations of increments of the multifractal random walk
    process.}
    \label{fig:synthetic_clean_samples}
\end{figure*}

Figure \ref{fig:synthetic_clean_samples} shows realizations of the multifractal random walk process used
in the stylized example.

\begin{figure*}[!h]
    \centering
    \subfloat{\includegraphics[width=0.495\linewidth]
    {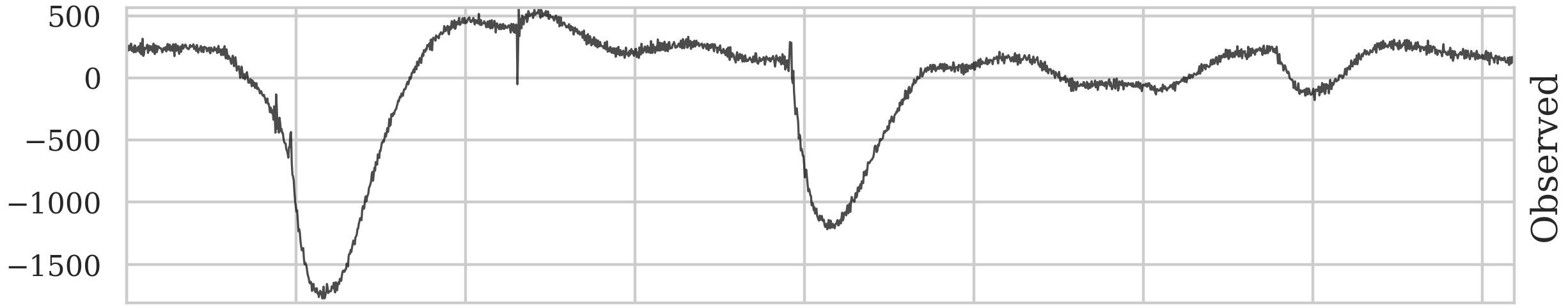}}
    \subfloat{\includegraphics[width=0.495\linewidth]
    {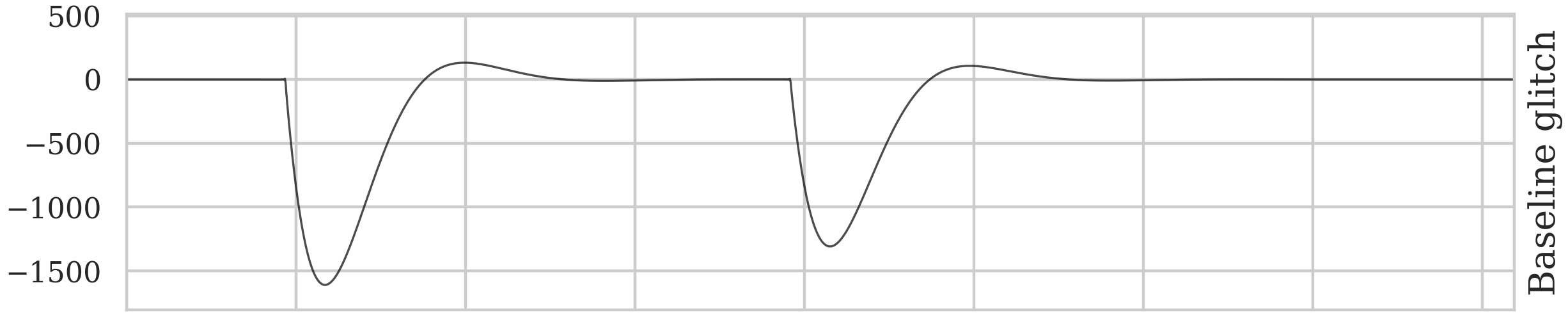}}
    \\
    \subfloat{\includegraphics[width=0.495\linewidth]
    {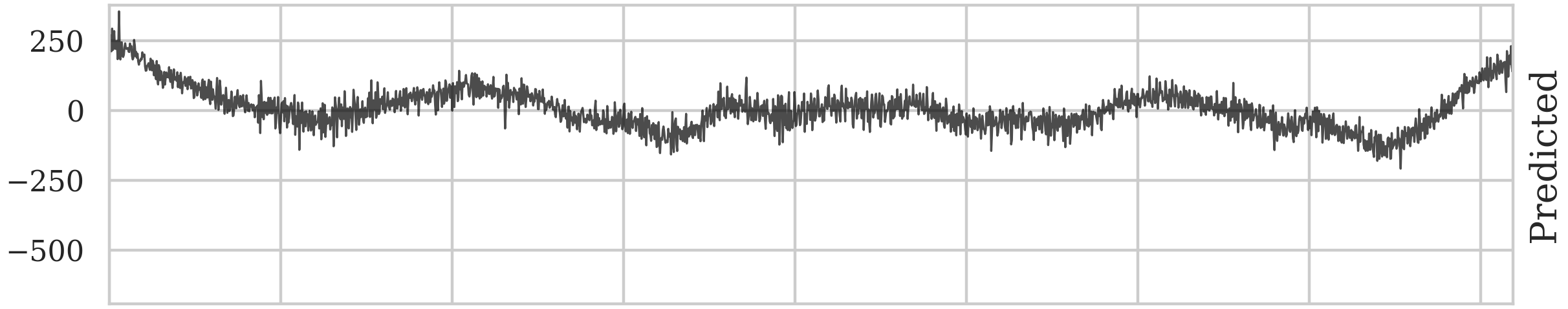}}
    \subfloat{\includegraphics[width=0.495\linewidth]
    {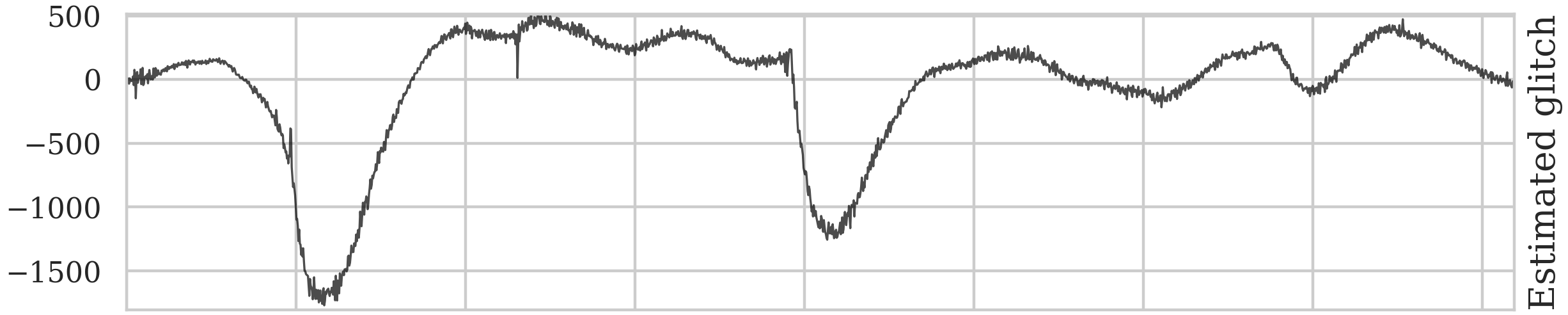}}
    \caption{Unsupervised source separation for glitch removal.}
    \label{fig:mars_example_2}
\end{figure*}

\begin{figure*}[!h]
    \centering
    \subfloat{\includegraphics[width=0.495\linewidth]
    {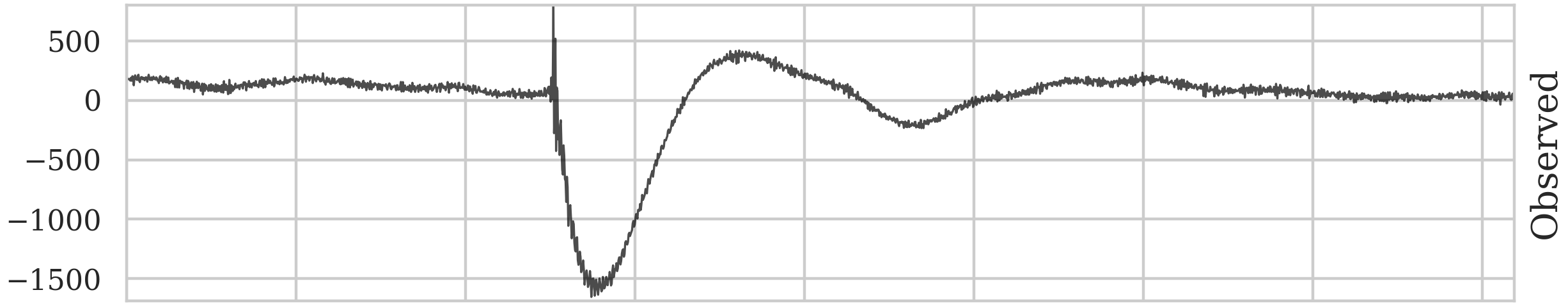}}
    \subfloat{\includegraphics[width=0.495\linewidth]
    {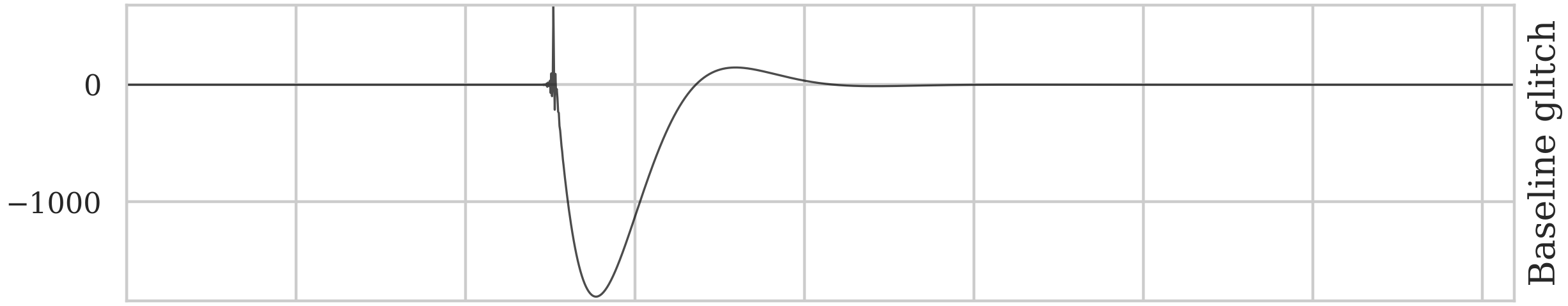}}
    \\
    \subfloat{\includegraphics[width=0.495\linewidth]
    {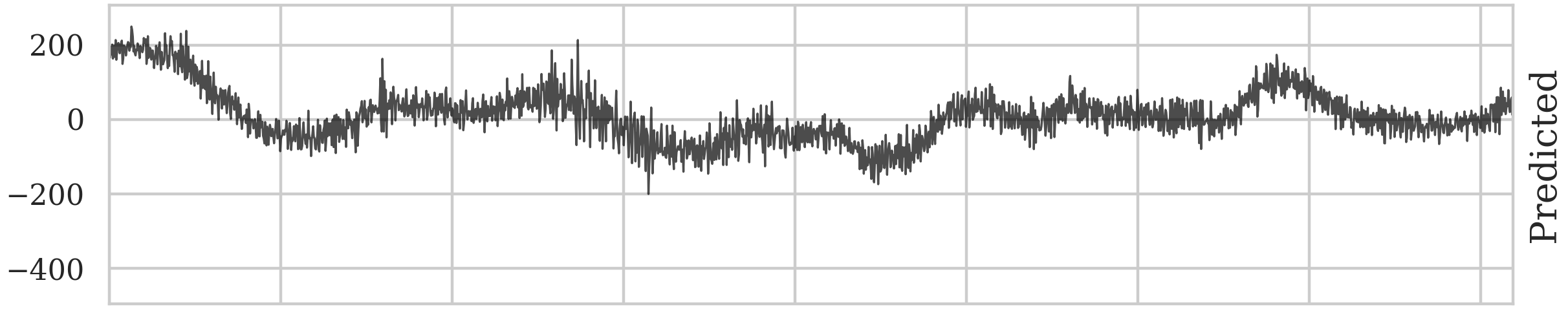}}
    \subfloat{\includegraphics[width=0.495\linewidth]
    {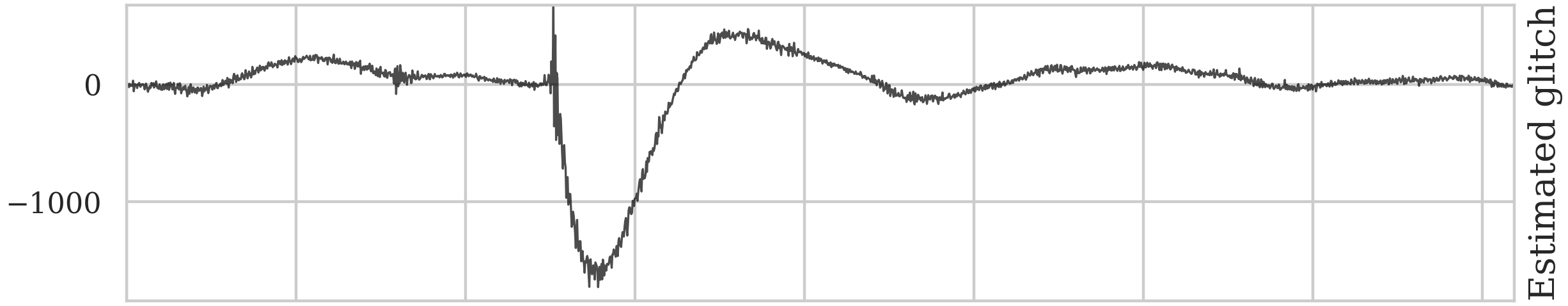}}
    \caption{Unsupervised source separation for glitch removal.}
    \label{fig:mars_example_3}
\end{figure*}

\begin{figure*}[!h]
    \centering
    \subfloat{\includegraphics[width=0.495\linewidth]
    {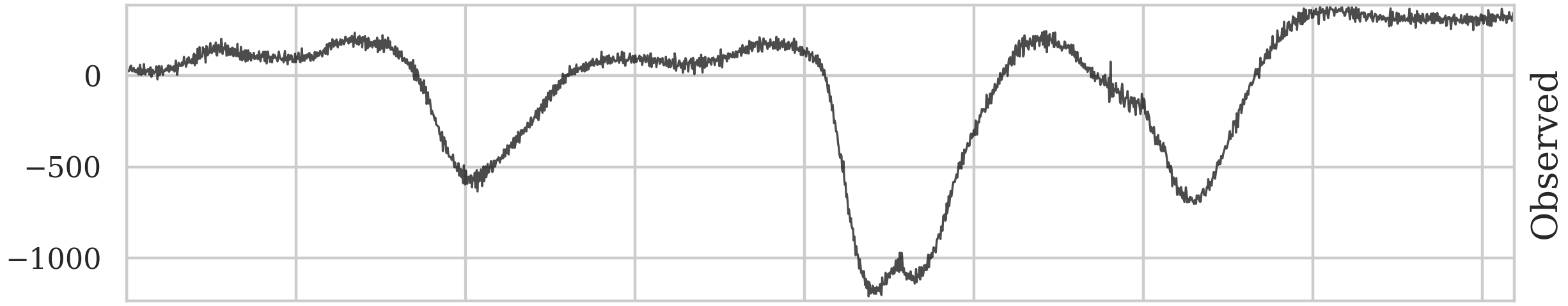}}
    \subfloat{\includegraphics[width=0.495\linewidth]
    {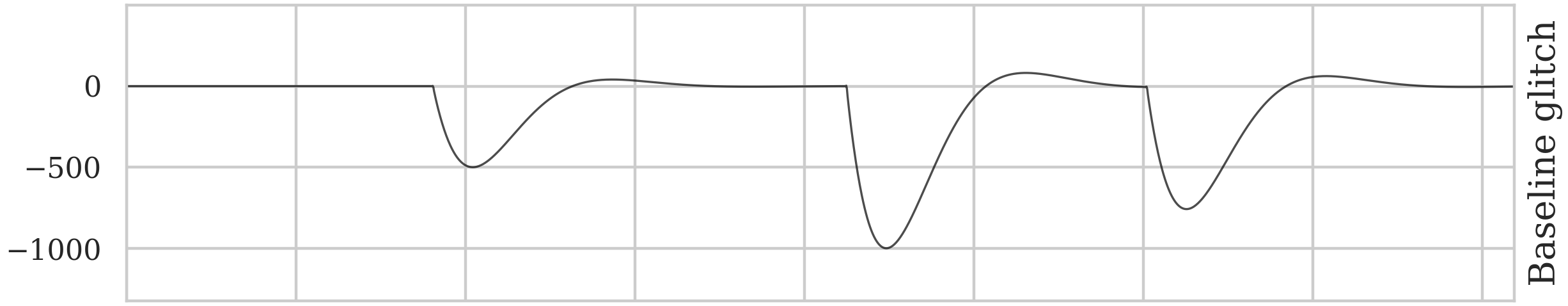}}
    \\
    \subfloat{\includegraphics[width=0.495\linewidth]
    {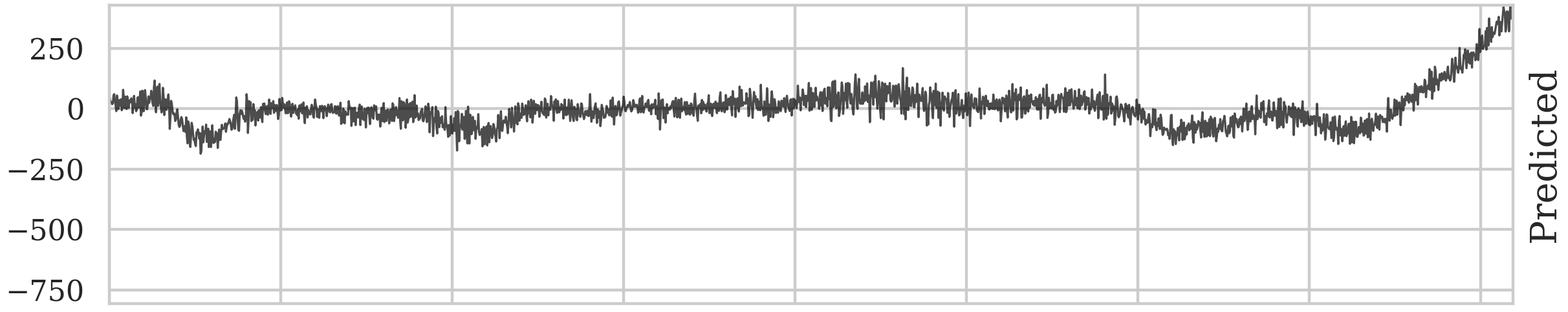}}
    \subfloat{\includegraphics[width=0.495\linewidth]
    {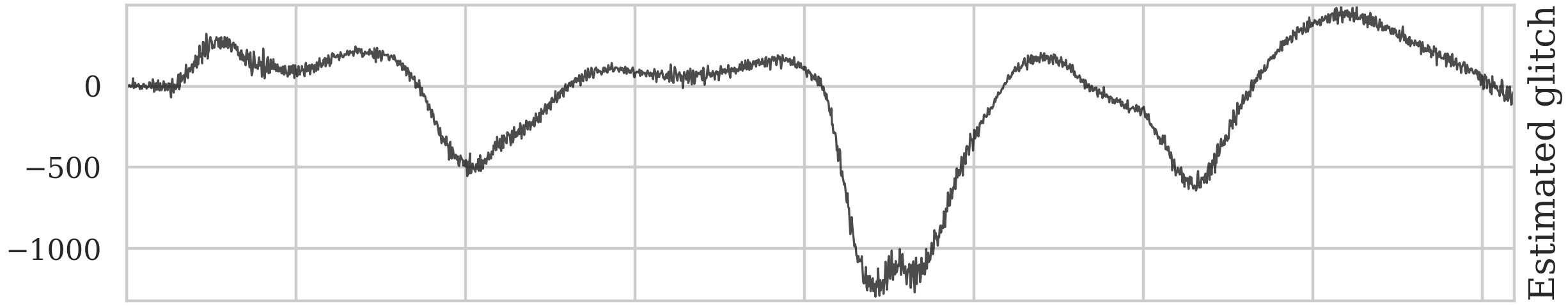}}
    \caption{Unsupervised source separation for glitch removal.}
    \label{fig:mars_example_4}
\end{figure*}

\begin{figure*}[!h]
    \centering
    \subfloat{\includegraphics[width=0.495\linewidth]
    {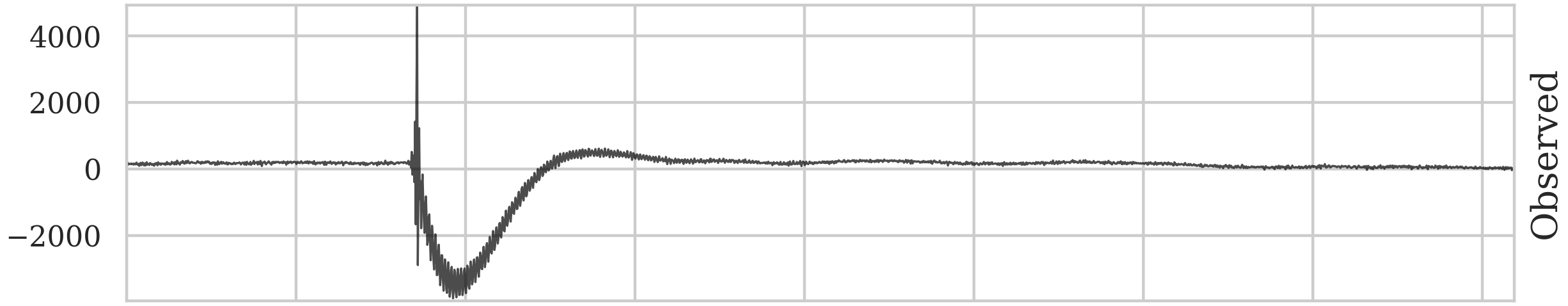}}
    \subfloat{\includegraphics[width=0.495\linewidth]
    {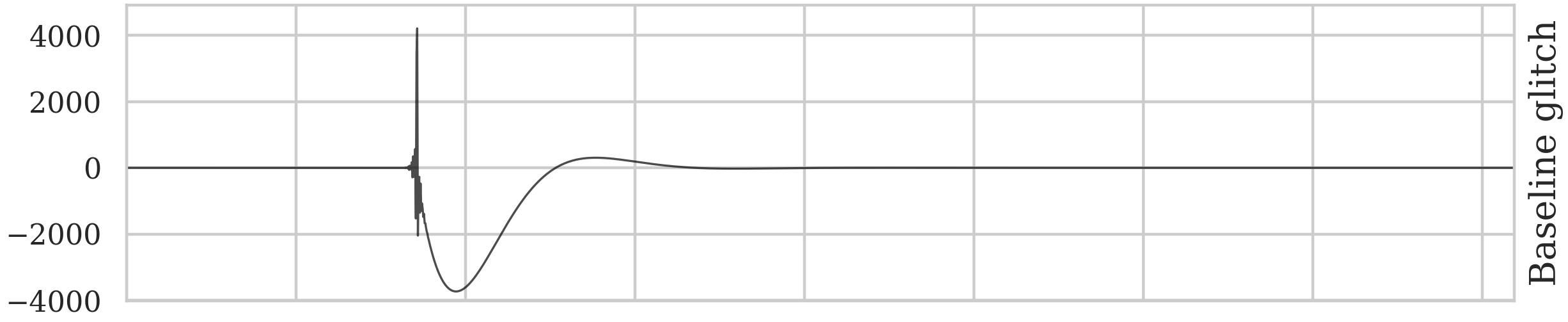}}
    \\
    \subfloat{\includegraphics[width=0.495\linewidth]
    {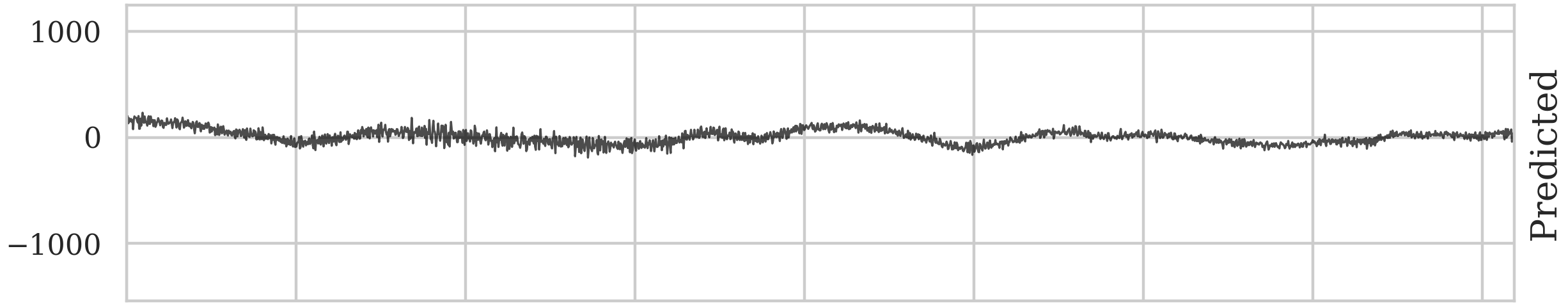}}
    \subfloat{\includegraphics[width=0.495\linewidth]
    {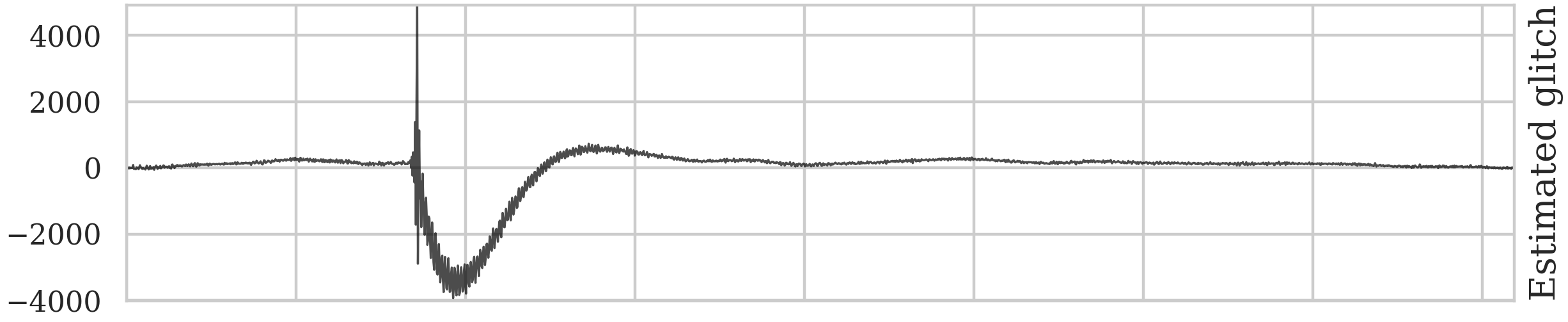}}
    \caption{Unsupervised source separation for glitch removal.}
    \label{fig:mars_example_5}
\end{figure*}

\section{Additional Glitch Separation Results}
\label{additional_results}

Here we provide more results regarding separating glitches from the seismic data recorded during the NASA InSight mission. Figures \ref{fig:mars_example_2}--\ref{fig:mars_example_5} provide glitch removal results for a more diverse set of glitches using the same setup as described in section~\ref{sec:removing_glitches_main}.

We provide more comprehensive deglitching results by applying our approach to perform glitch separation on the U
component for the nighttime (17:08--00:55 LMST) during sol 187 (June 6, 2019), as
the glitches during the day are often obscured by daytime noise. We used a set
of 50 snippets with window size of $204.8\,\mathrm{s}$ and solved the source
separation optimization problem using $200$ L-BFGS iterations.

\begin{figure}[!t]
    \centering

    \begin{subfigure}[b]{1.0\linewidth}
        \includegraphics[width=1.0\linewidth]{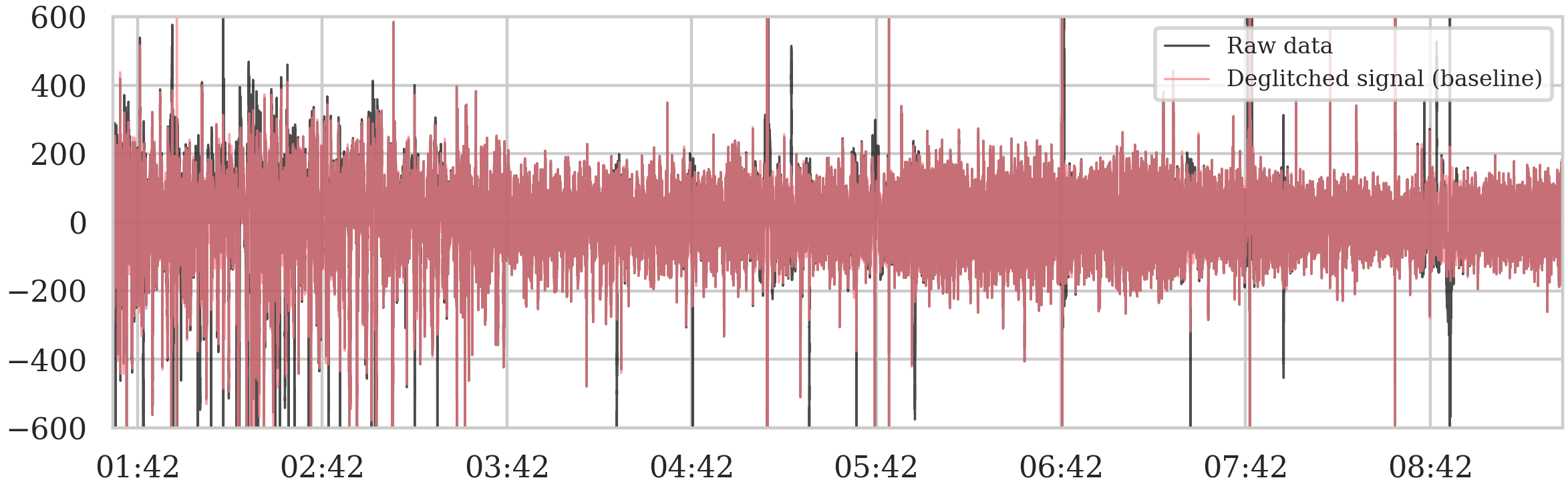}
    \end{subfigure}\hspace{0em}
    \label{Basline}

    \begin{subfigure}[b]{1.0\linewidth}
        \includegraphics[width=1.0\linewidth]{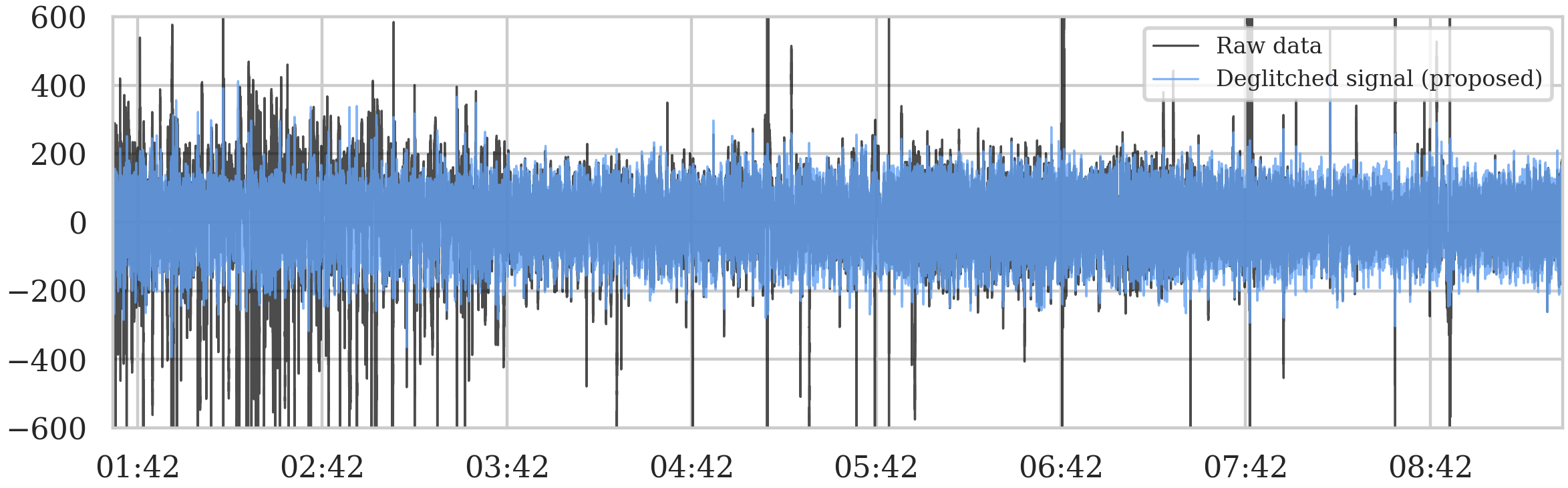}
    \end{subfigure}\hspace{0em}
    \label{Proposed method}

    \caption{Unsupervised separation of glitches from seismic data recorded during sol 187 (June 6, 2019) from 17:08 to 00:55 Martian local time (the horizontal axis is in UTC time zone). The raw data is depicted in black, with the predicted deglitched data overlaid, represented by the baseline method in red and the proposed method in blue. The high-amplitude ``spikes'' observed in the raw waveform correspond to glitches. A successful deglitching outcome should exclude these spikes. Our deglitching results effectively separate a significant number of these high-amplitude events, whereas the baseline method fails to address a considerable portion of them.}
    \label{fig:sol_187_deglitching}
\end{figure}

\begin{figure}[!t]
    \centering

    \begin{subfigure}[b]{1.0\linewidth}
        \includegraphics[width=1.0\linewidth]{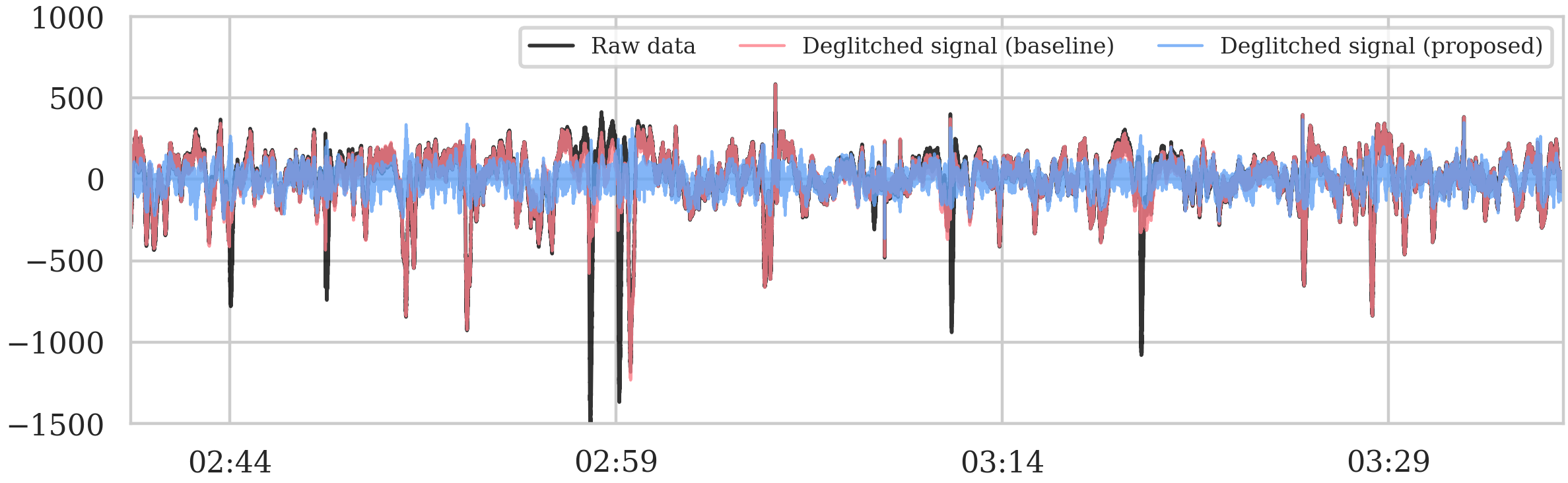}
    \end{subfigure}\hspace{0em}

    \begin{subfigure}[b]{1.0\linewidth}
        \includegraphics[width=1.0\linewidth]{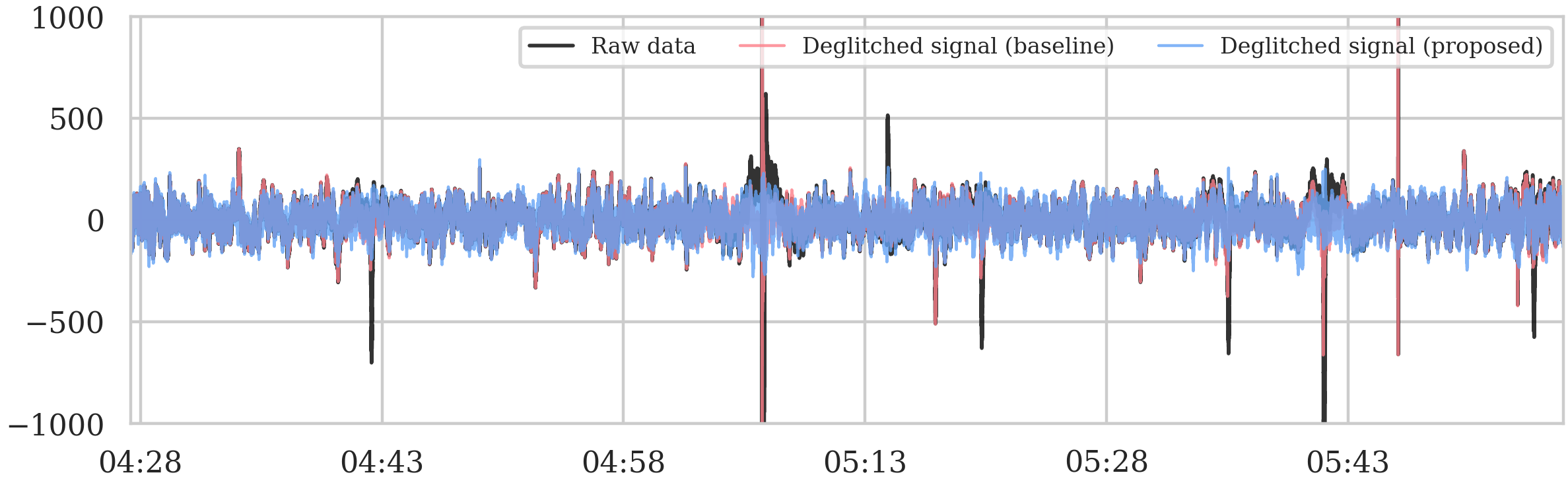}
    \end{subfigure}\hspace{0em}

    \begin{subfigure}[b]{1.0\linewidth}
        \includegraphics[width=1.0\linewidth]{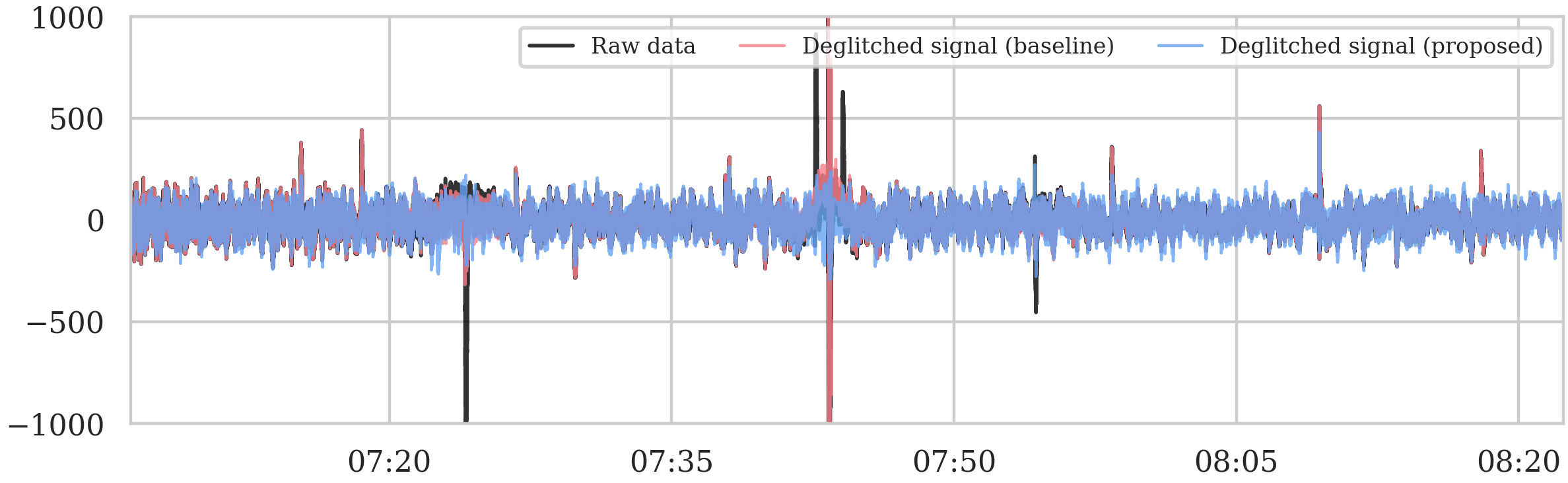}
    \end{subfigure}\hspace{0em}

    \caption{Three zoomed-in time intervals from Figure~\ref{fig:sol_187_deglitching} to facilitate a detailed performance comparison between the baseline (red) and the proposed deglitching results. Both outcomes are overlaid on the raw waveform shown in black. The glitches manifest as high-amplitude one-sided pulses in the raw waveform, which we intend to separate. Within each of the aforementioned time intervals, it is evident that the baseline approach falls short in effectively separating several glitches. The horizontal axis represents the UTC time zone.}
    \label{fig:sol_187_deglitching_zoom}
\end{figure}

Our results indicate that the baseline method appears to overlook
several anomalies in the U component that we believe to be glitches. In
contrast, our method not only detects all the glitches identified by the
baseline method, but it also recognizes a significant number of additional
glitches. Although it is true that our method appears to detect more glitches
than the baseline, we must recognize that the baseline is the only dependable
reference for identifying glitches and further verification by
InSight experts is necessary to confirm the legitimacy of the identified events
as glitches.

\section{Additional Marsquake Background Noise Separation Results}
\label{additional_marsquake_results}

We present additional results on the separation of marsquake background noise and glitches, showcasing different marsquake characteristics. The first example pertains to a marsquake recorded on January 2, 2022~\cite{doi.org/10.12686/a19}. This particular marsquake exhibits a larger amplitude and a longer coda wave compared to the one presented in Figure~\ref{fig:marsquake_cleaning}. Although the background noise appears negligible and is not readily visible in the raw waveform, this provides an opportunity to demonstrate the effectiveness of our unsupervised source separation method when one source (the marsquake in this case) dominates in amplitude.

\begin{figure}[t]
    \centering

    \begin{subfigure}[b]{1.0\linewidth}
        \includegraphics[width=1.0\linewidth]{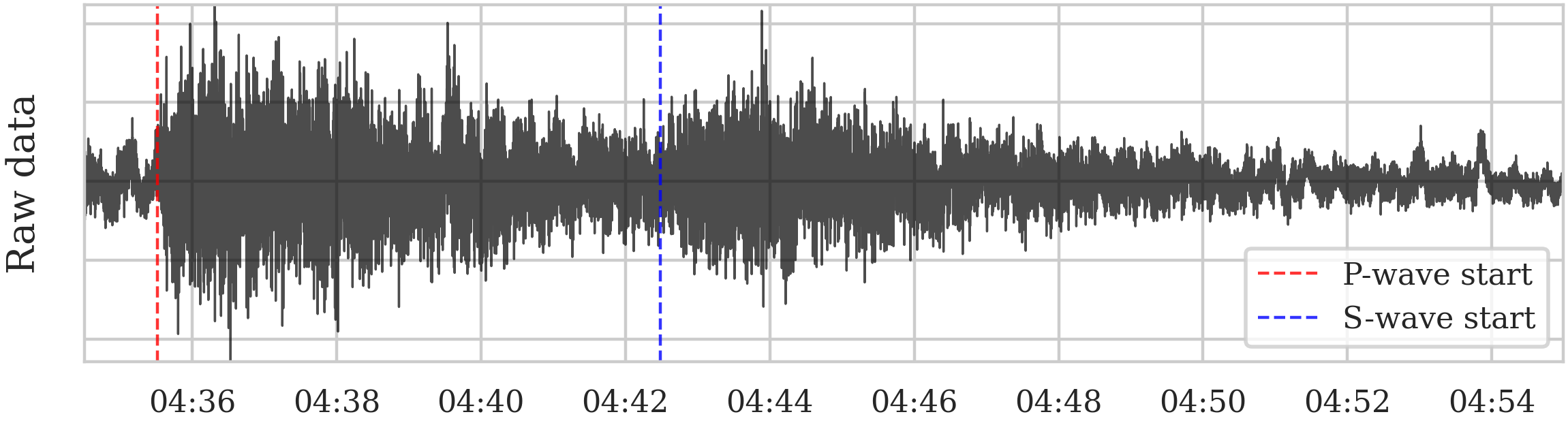}
    \end{subfigure}\hspace{0em}

    \begin{subfigure}[b]{1.0\linewidth}
        \includegraphics[width=1.0\linewidth]{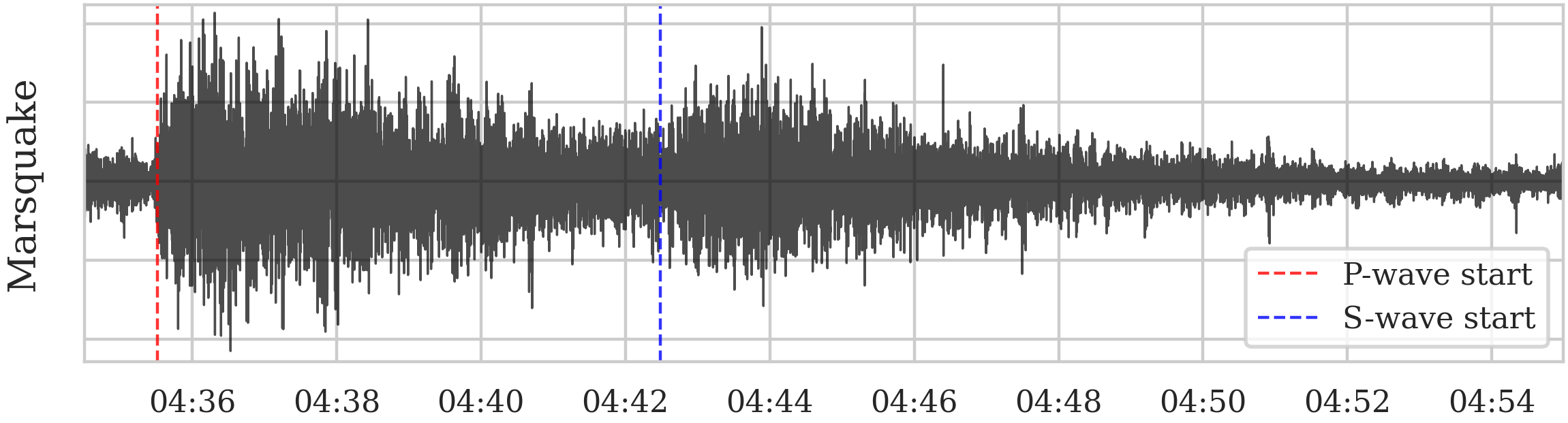}
    \end{subfigure}\hspace{0em}

    \begin{subfigure}[b]{1.0\linewidth}
        \includegraphics[width=1.0\linewidth]{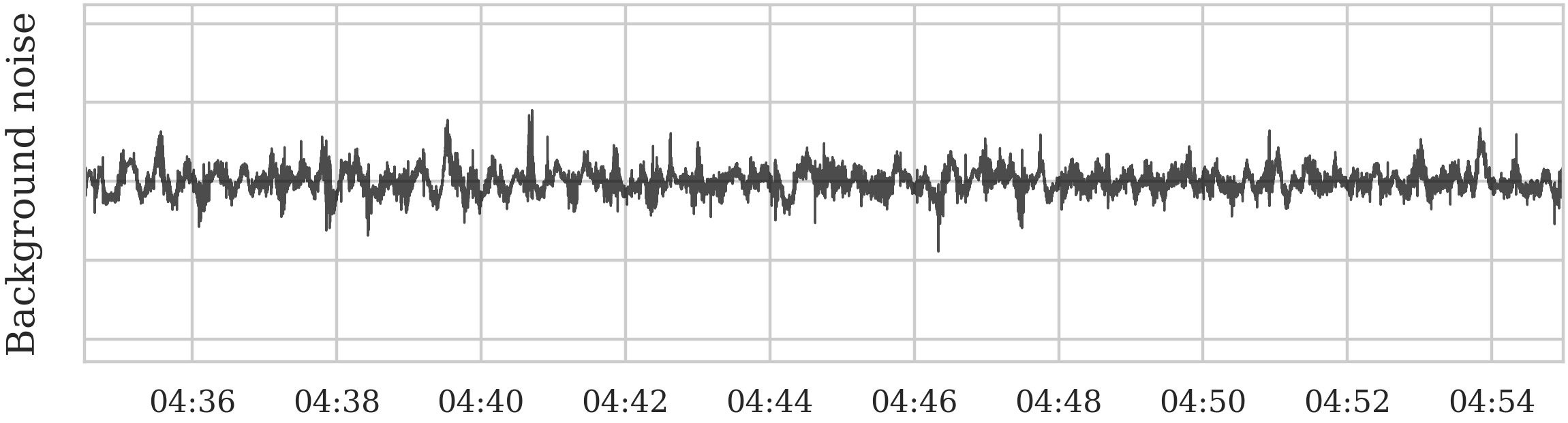}
    \end{subfigure}\hspace{0em}
    \caption{Unsupervised separation of background noise and glitches from a marsquake recorded by the InSight lander's seismometer on January 2, 2022~\cite{doi.org/10.12686/a19}. Approximately 36 hours of raw data from the U component were used without any additional prior knowledge of marsquakes or glitches. The horizontal axis is in UTC time zone.}
    \label{fig:marsquake_cleaning_S1102a}
\end{figure}

To achieve the separation of background noise, we selected approximately 36 hours of detrended raw data from the U component with a sampling rate of 20Hz. This ensured an accurate estimation of the wavelet scattering covariance statistics. The network architecture used is the same as in previous examples, and we employed a window size of $204.8,$s. By solving the optimization problem outlined in equation~\eqref{total_loss} with 200 L-BFGS iterations, we obtained the results depicted in Figure~\ref{fig:marsquake_cleaning_S1102a}. Notably, glitches occurring just before the P-wave arrival and towards the end of the marsquake were successfully separated. Moreover, the separated background noise exhibits a stationary characteristic, which is desirable as it indicates minimal leakage of the marsquake signal.

\begin{figure}[t]
    \centering

    \begin{subfigure}{1.0\linewidth}
        \includegraphics[width=1.0\linewidth]{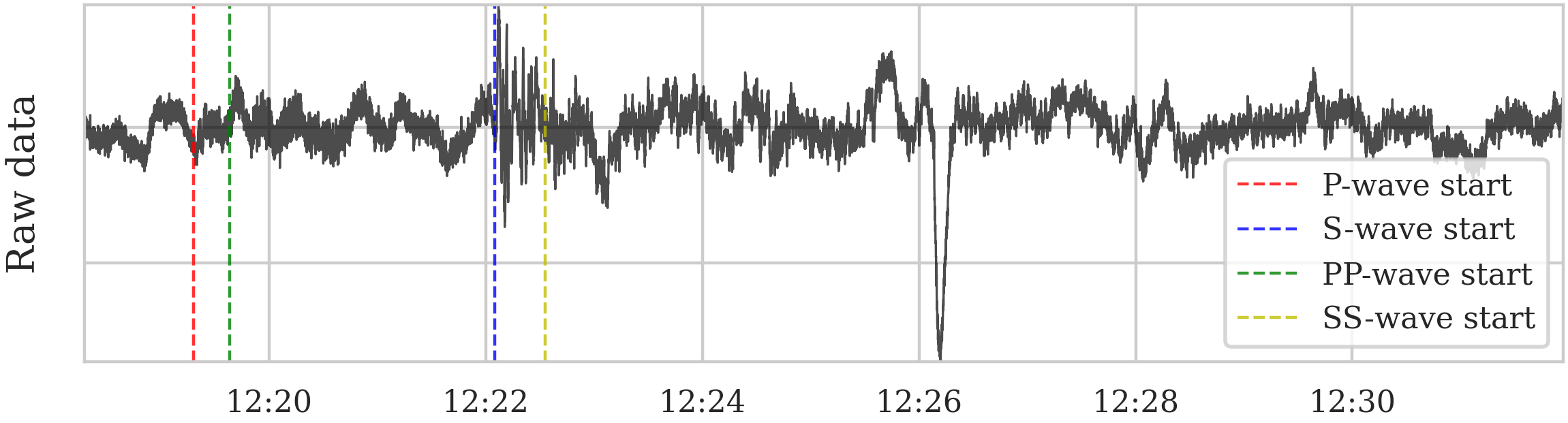}
    \end{subfigure}\hspace{0em}

    \begin{subfigure}{1.0\linewidth}
        \includegraphics[width=1.0\linewidth]{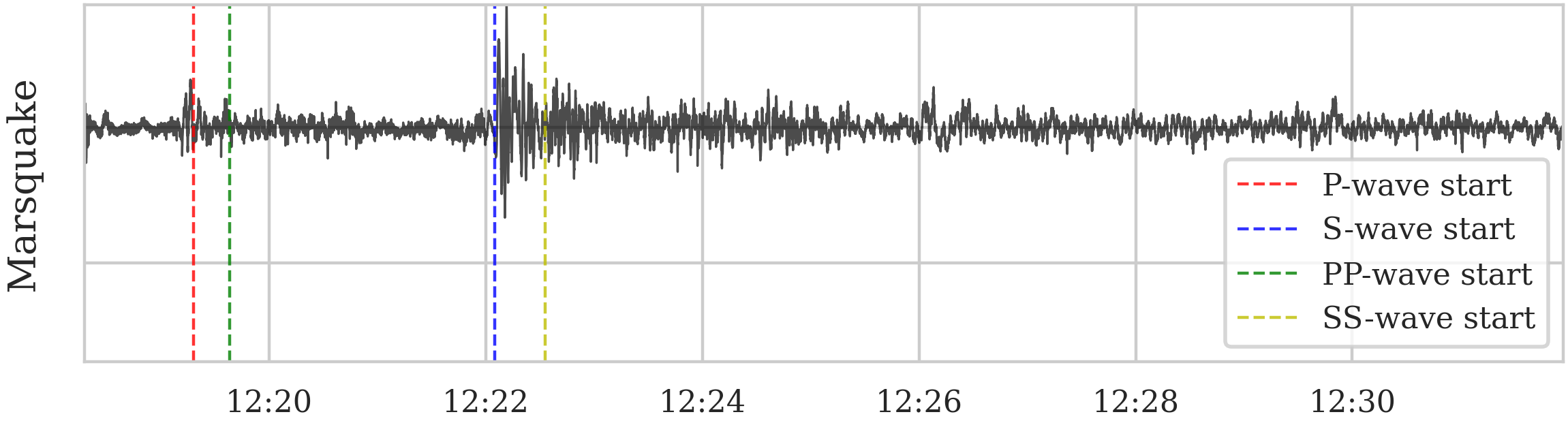}
    \end{subfigure}\hspace{0em}

    \begin{subfigure}{1.0\linewidth}
        \includegraphics[width=1.0\linewidth]{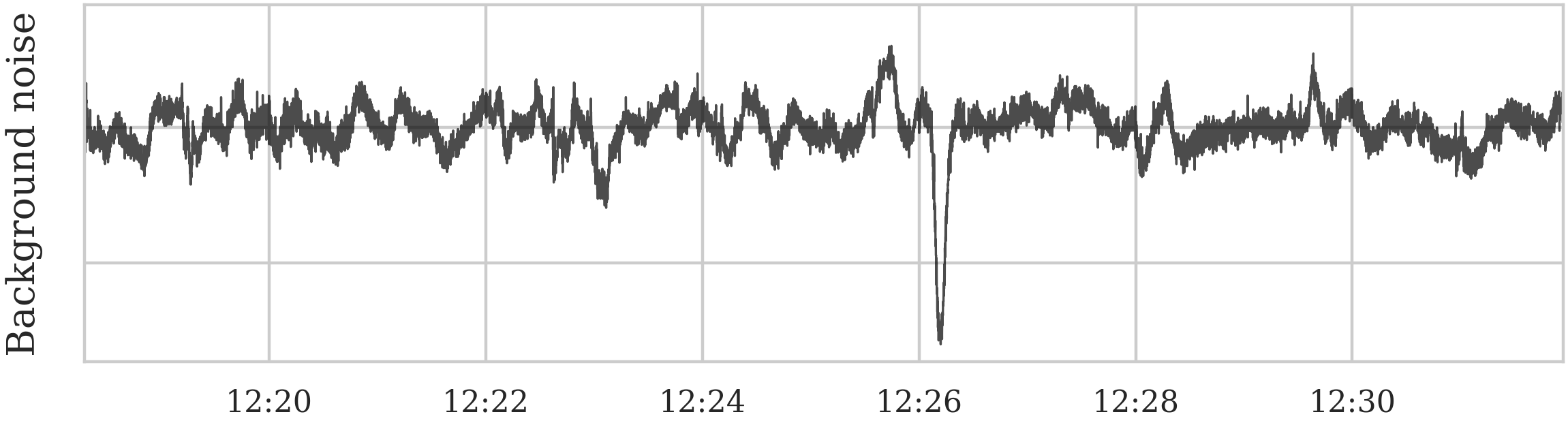}
    \end{subfigure}\hspace{0em}
    \caption{Unsupervised separation of background noise and glitches from a marsquake recorded by the InSight lander's seismometer on July 26, 2019~\cite{doi.org/10.12686/a19}. Approximately 95 hours of raw data from the U component were used without any additional prior knowledge of marsquakes or glitches. The horizontal axis is in UTC time zone.}
    \label{fig:marsquake_cleaning_S0235b}
\end{figure}

The final example involves a marsquake recorded on July 26, 2019~\cite{doi.org/10.12686/a19}. Separating the background noise in this case proves more challenging, as the P-wave arrival is barely discernible in the raw waveform shown in the top panel of Figure~\ref{fig:marsquake_cleaning_S0235b}. Furthermore, the presence of background noise masks the detection of the S-wave, as well as the secondary PP- and SS-wave arrivals. To address these complexities and achieve accurate separation of the marsquake while minimizing signal leakage, we require 95 hours of detrended raw data from the U component. A window size of $409.6\,$s is used, and the optimization problem in equation~\eqref{total_loss} is solved with 200 L-BFGS iterations. The results are depicted in Figure~\ref{fig:marsquake_cleaning_S0235b}, where the separated marsquake is distinctly delineated. The accuracy of our approach is further confirmed by the independently picked arrival times by the InSight team~\cite{10.1029/2020EA001317}, shown as dotted lines in Figure~\ref{fig:marsquake_cleaning_S0235b}. The alignment between their picked arrival times and our separated marsquake serves as validation for the accuracy of our method.

\newpage
\null

\end{document}